\documentclass[3p,,preprint,12pt]{elsarticle}

\usepackage{filecontents}

\makeatletter\if@twocolumn\PassOptionsToPackage{switch}{lineno}\else\fi\makeatother

\usepackage{tabulary,xcolor}
\usepackage{amsfonts,amsmath,amssymb}
\usepackage[T1]{fontenc}
\makeatletter
\let\save@ps@pprintTitle\ps@pprintTitle
\def\ps@pprintTitle{\save@ps@pprintTitle\gdef\@oddfoot{\footnotesize\itshape \null\hfill\today}}
\def\hlinewd#1{%
  \noalign{\ifnum0=`}\fi\hrule \@height #1%
  \futurelet\reserved@a\@xhline}
\def\tbltoprule{\hlinewd{.8pt}\\[-12pt]}
\def\tblbottomrule{\noalign{\vspace*{6pt}}\hline\noalign{\vspace*{2pt}}}

\AtBeginDocument{\ifNAT@numbers \biboptions{sort&compress}\fi}
\makeatother

\usepackage[colorlinks=true]{hyperref}
\usepackage{caption}
\usepackage{subcaption}

\usepackage{ifluatex}
\ifluatex
\usepackage{fontspec}
\defaultfontfeatures{Ligatures=TeX}
\usepackage[]{unicode-math}
\unimathsetup{math-style=TeX}
\else 
\usepackage[utf8]{inputenc}
\fi 
\ifluatex\else\usepackage{stmaryrd}\fi

\usepackage{pdflscape}  
\usepackage{url,multirow,morefloats,floatflt,cancel,tfrupee}
\makeatletter

\AtBeginDocument{\@ifpackageloaded{textcomp}{}{\usepackage{textcomp}}}
\makeatother
\usepackage{colortbl}
\usepackage{xcolor}
\usepackage{pifont}
\usepackage{float}
\usepackage[nointegrals]{wasysym}
\usepackage{setspace}
\usepackage[section]{placeins}
\makeatletter

\def\mcWidth#1{\csname TY@F#1\endcsname+\tabcolsep}

\def\cAlignHack{\rightskip\@flushglue\leftskip\@flushglue\parindent\z@\parfillskip\z@skip}
\def\rAlignHack{\rightskip\z@skip\leftskip\@flushglue \parindent\z@\parfillskip\z@skip}

\@ifundefined{etal}{}{}

\usepackage{ifxetex}
\ifxetex\else\if@twocolumn\@ifpackageloaded{stfloats}{}{\usepackage{dblfloatfix}}\fi\fi

\AtBeginDocument{
\expandafter\ifx\csname eqalign\endcsname\relax
\def\eqalign#1{\null\vcenter{\def\\{\cr}\openup\jot\m@th
  \ialign{\strut$\displaystyle{##}$\hfil&$\displaystyle{{}##}$\hfil
      \crcr#1\crcr}}\,}
\fi
}

\AtBeginDocument{%
  \@ifpackageloaded{endfloat}%
   {\renewcommand\efloat@iwrite[1]{\immediate\expandafter\protected@write\csname efloat@post#1\endcsname{}}}{\newif\ifefloat@tables}%
}%

\def\BreakURLText#1{\@tfor\brk@tempa:=#1\do{\brk@tempa\hskip0pt}}
\let\lt=<
\let\gt=>
\def\processVert{\ifmmode|\else\textbar\fi}

\@ifundefined{subparagraph}{
\def\subparagraph{\@startsection{paragraph}{5}{2\parindent}{0ex plus 0.1ex minus 0.1ex}%
{0ex}{\normalfont\small\itshape}}%
}{}

\newcommand\role[1]{\unskip}
\newcommand\aucollab[1]{\unskip}
  
\@ifundefined{tsGraphicsScaleX}{\gdef\tsGraphicsScaleX{1}}{}
\@ifundefined{tsGraphicsScaleY}{\gdef\tsGraphicsScaleY{.9}}{}
\def\checkGraphicsWidth{\ifdim\Gin@nat@width>\linewidth
	\tsGraphicsScaleX\linewidth\else\Gin@nat@width\fi}

\def\checkGraphicsHeight{\ifdim\Gin@nat@height>.9\textheight
	\tsGraphicsScaleY\textheight\else\Gin@nat@height\fi}

\def\fixFloatSize#1{}
\let\ts@includegraphics\includegraphics

\def\inlinegraphic[#1]#2{{\edef\@tempa{#1}\edef\baseline@shift{\ifx\@tempa\@empty0\else#1\fi}\edef\tempZ{\the\numexpr(\numexpr(\baseline@shift*\f@size/100))}\protect\raisebox{\tempZ pt}{\ts@includegraphics{#2}}}}

\AtBeginDocument{\def\includegraphics{\@ifnextchar[{\ts@includegraphics}{\ts@includegraphics[width=\checkGraphicsWidth,height=\checkGraphicsHeight,keepaspectratio]}}}

\DeclareMathAlphabet{\mathpzc}{OT1}{pzc}{m}{it}

\def\URL#1#2{\@ifundefined{href}{#2}{\href{#1}{#2}}}

\def\UrlOrds{\do\*\do\-\do\~\do\'\do\"\do\-}%
\g@addto@macro{\UrlBreaks}{\UrlOrds}

\edef\fntEncoding{\f@encoding}

\makeatother

\newif\ifmultipleabstract\multipleabstractfalse%
\def\truncatedAt{1000}
\def\typesetArtId{722523e1-9528-421e-baf5-8185fa164563}    
    \makeatletter
    \@ifpackageloaded{hyperref}{}{\usepackage[pdfborder={0 0 0},bookmarks=false]{hyperref}}
    \@ifpackageloaded{hyperref}{}{}%
    \def\putUpgradeInfoBox{\@ifundefined{truncatedAt}{\def\truncatedAt{1000}}{}
    \def\up@width@one{\if@twocolumn .95\columnwidth\else .65\columnwidth\fi}%
    \def\up@width@two{\if@twocolumn .5\columnwidth\else .3\columnwidth\fi}%
    \vskip 2pc\nopagebreak
    \noindent\vbox{\centering%
    {\fontfamily{phv}\selectfont\footnotesize\color{blue}%
    	\ifx\typesetArtId\empty%
      	\href{https://typeset.io/documents}{\underline{Edit this article on \smash{Typeset}}}%
      \else%
	    	\href{https://typeset.io/edit/\typesetArtId}{\underline{Edit this article on \smash{Typeset}}}%
      \fi%
      }\\[6pt]
    \par\href{https://www.typeset.io/pricing/?source=upgrade-from-pdf}{\includegraphics[width=\up@width@one]{upgrade-logo-11Nov19.png}}\\[2pt]%
    \par%
    \fontfamily{phv}\selectfont\footnotesize\color{blue}%
    \href{https://typeset.io/}{\underline{www.\smash{typeset}.io}}%
    ~~\,\textcolor{black}{\vrule height 8pt width .7pt}~\,~%
    \href{https://www.typeset.io/orders/coupon/ZfAB8STU/?source=pricing-page-student-discount}{\underline{\smash{Looking for a Discount?}}}%
    }%
    }
    \makeatother

\begin{document}

\begin{frontmatter}
	

\title{
    Human-robot collaboration and machine learning: a systematic review of recent research 
}

\address[manch]{
Cognitive Robotics Laboratory, The University of Manchester, Manchester, United Kingdom
}
\address[bae]{
BAE Systems (Operations) Ltd., Warton Aerodrome, Lancashire, United Kingdom
}
    
\author[manch]{Francesco Semeraro\corref{cor1}}
\ead{francesco.semeraro@manchester.ac.uk}
\author[bae]{Alexander Griffiths}
\ead{alexander.griffiths@baesystems.com}
\author[manch]{Angelo Cangelosi}
\ead{angelo.cangelosi@manchester.ac.uk}

\cortext[cor1]{Corresponding author}

\begin{abstract}
Technological progress increasingly envisions the use of robots interacting with people in everyday life. Human-robot collaboration (HRC) is the approach that explores the interaction between a human and a robot, during the completion of a common objective, at the cognitive and physical level. In HRC works, a cognitive model is typically built, which collects inputs from the environment and from the user, elaborates and translates these into information that can be used by the robot itself. Machine learning is a recent approach to build the cognitive model and behavioural block, with high potential in HRC. Consequently, this paper proposes a thorough literature review of the use of machine learning techniques in the context of human-robot collaboration. 45 key papers were selected and analysed, and a clustering of works based on the type of collaborative tasks, evaluation metrics and cognitive variables modelled is proposed. Then, a deep analysis on different families of machine learning algorithms and their properties, along with the sensing modalities used, is carried out. Among the observations, it is outlined the importance of the machine learning algorithms to incorporate time dependencies. The salient features of these works are then cross-analysed to show trends in HRC and give guidelines for future works, comparing them with other aspects of HRC not appeared in the review.  

\section*{Highlights}
\begin{itemize}
   \item Machine learning is progressively assuming an important role in human-robot collaboration.
   \item Categorizations of human-robot collaboration experiments based on the type of tasks, evaluation metrics, cognitive variables, machine learning techniques and sensing modalities are proposed.
   \item It is crucial to make machine learning algorithms sensitive to time dependencies.
   \item The use of composite machine learning models for cognitive systems, among which deep reinforcement learning, should be encouraged.
   \item 
   Use of different architectures than robotic manipulators, digital twin and non-dyadic interactions, are potential research paths to explore.
\end{itemize}

\begin{keyword}
Human-robot collaboration \sep Collaborative robotics \sep Cobot \sep Machine learning \sep Human-robot interaction
\end{keyword}

\end{abstract}

\end{frontmatter}

\section{Introduction}
\label{introduction}
Robots are progressively occupying a relevant place in everyday life \unskip~\cite{IFR2019}. Technological progress increasingly envisions the use of robots interacting with people in numerous application domains such as flexible manufacturing scenarios, assistive robotics, social companions. Therefore, human-robot interaction studies are crucial for the proper deployment of robots at a closer contact with humans. Among the possible declinations of this sector of robotics research, human-robot collaboration (HRC) is the branch that explores the interaction between a human and a robot acting like a team in reaching a common goal \unskip~\cite{bauer2008human}. Majority of works of this kind focus on such a team collaboratively working on an actual physical task, at the cognitive and physical level. In these HRC works, a cognitive model (or cognitive architecture) is typically built, which collects inputs from the environment and from the user, elaborates and translates them into information that can be used by the robot itself. The robot will use this model to tune its own behaviour, to increase the joint performance with the user.

Studies in HRC specifically explore the feasibility of use of robotic platforms at a very close proximity with the user for joint tasks. This sector mainly depicts the use of robotic manipulators that share the same workspace and jointly work at the realisation of a piece of manufacturing. Such application of robotics is called cobotics and robots that can safely serve this purpose gain the denomination of collaborative robots or CoBots (from this point on, they will be referred as "cobots" in this work) \unskip~\cite{IFR2019}. 

The objective of cobots is the increase of the performance of the manufacturing process, both in terms of productivity yield and of relieving the user from the burden of the task, either physically or cognitively \unskip~\cite{Matheson2019}. Besides, there are also other sectors that could benefit from the use of collaborative robots. Indeed, HRC has started to be considered in surgical scenarios \unskip~\cite{ISI:000389809100162}. In these, robotic manipulators are foreseen helping surgeons, by passing tools \unskip~\cite{jacob2012gestonurse} or by assisting them in the execution of an actual surgical procedure \unskip~\cite{amarillo2021collaborative}.

Such an interest for this sector of robotics research has led to the publications of hundreds of works on HRC \unskip~\cite{Ogata2022}. Consequently, recent attempts to regularise and standardise the knowledge about this topic, through reviews and surveys, have already been pursued. As mentioned, in Matheson et al. \unskip~\cite{Matheson2019}, a pool of works about the use of human-robot collaborations in manufacturing applications was analysed. They highlight the distinction between works that considered an industrial robot and a cobot \unskip~\cite{IFR2019}. The main difference between these is that the design of a cobot and its cognitive architecture makes it more suitable to be at closer reach with humans than the industrial one.

Secondly, in Ogenyi et al. \unskip~\cite{EmeohaOgenyi} a survey about human-robot collaboration was pursued, providing a classification looking at aspects related to the collaborative strategies, learning methods, sensors and actuators. However, they specifically focused on physical human-robot collaboration, without taking the cognitive aspect of such interaction into account.

Interestingly, instead of addressing the robotic side of the interaction, Hiatt et al. \unskip~\cite{Hiatt2017} focused on the human one. They proposed an attempt of standardization of the human modelling in human-robot collaboration, both from a mental and computational point of view. Chandrasekaran et al. \unskip~\cite{Chandrasekaran2015} produced an overview of human-robot collaboration which concentrated on the sectors of applications themselves. Finally, more recently, Mukherjee et al. \unskip~\cite{mukherjee2022survey} provided a thorough review about the design of learning strategies in HRC.

To the best of our knowledge, there is no previous attempt of producing a review with a special focus on the contributions of machine learning (ML) to such a sector. Machine learning is a promising field of computer science that trains systems, by means of algorithms and learning methods, to recognise and classifies instances of data never experienced by the system before. This description is explanatory of how much ML can empower the sector of human-robot collaboration, by embedding itself in the behavioural block of the robotic system \unskip~\cite{IFR2019,EmeohaOgenyi}. Therefore, this paper provides a thorough review on the use of machine learning techniques in human-robot collaboration research. For a more comprehensive understanding of the use of ML in HRC, this review also analyses the types of experiment carried out in this sector of robotics, proposing a categorization which is independent from the applications which motivated the works. Furthermore, the review also addresses other aspects related to the experimental design of HRC, by looking at the sensing used and variables modelled through the use of machine learning algorithms. 

The paper is structured as follows. Section \ref{selection} provides a full breakdown of the methodology used to perform the publication database research. Section \ref{experiments} describes the types of interaction surveyed and the evaluation metrics reported. Section \ref{variables} enlists the variables used to model the environment. Section \ref{ML} gives deeper insights about the machine learning algorithms employed in HRC. Section \ref{sensing} briefly discusses the sensing modalities exploited. Then, Section \ref{discussions} carries out further analysis on the extracted features, providing suggestions on designing HRC systems with machine learning, and mentions other potential HRC domains, not appeared in the review process, where machine learning could contribute to. Finally, Section \ref{conclusions} sums up the main observations resulting from the review.
    
\section{Research methodology and selection criteria}
\label{selection}
The research was performed in January 2021, using the following publication databases: IEEE Xplore, ISI Web of Knowledge and Scopus. The ensuing combination of keywords was used: (``human-robot collaborat*'' OR ``human-robot cooperat*'' OR ``collaborative robot*'' OR ``cobot*'' OR ``hrc'') AND ``learning''. This ensemble was chosen to make sure to cover most of the terms used to address a human-robot collaboration research, without, at the same time, being too selective on the type of machine learning algorithm used. This set of keywords was searched in the title, abstract and keywords records of the journal articles and conference proceedings written in English, from 2015 to 2020. 

Inputting this set of search parameters returned a total of 389 results from ISI Web of Knowledge (191 articles, 198 proceedings), 178 from IEEE Xplore (48 articles, 130 proceedings) and 486 from Scopus (206 articles and 280 proceedings). These three sets were unified and their duplicates were removed, leading from an initial set of 1053 papers to a combined set of 627 works. The first round of selection was performed by looking at the presence of keywords related to machine learning and human-robot collaboration in the title or in the abstract, through scanning of the selected papers. Furthermore, the publications which focused on the topic, but did not report an actual experimental study were discarded. This process shrank the combined set to a second group of 226 papers.
At this point, a second round of selection was performed according to the following criterion. This set of papers was analysed to identify the presence of the following elements in the papers:\newline
\begin{enumerate}
    \item use of any machine learning algorithm as part of the behavioural block of a robotic system;\newline
    \item joint presence of a human and a robot in a real physical experimental validation;\newline
    \item physical interaction between user and robot in the experimental procedure, through direct contact or mediated through an object;\newline
    \item final tangible joint goal to achieve in the experiment;\newline
    \item actual behavioural change of the robot during the experiment, upon extraction of a new piece of information through the use of the machine learning algorithm\newline
\end{enumerate}

\begin{figure}[H]
  \centering
  \includegraphics[width=\textwidth]{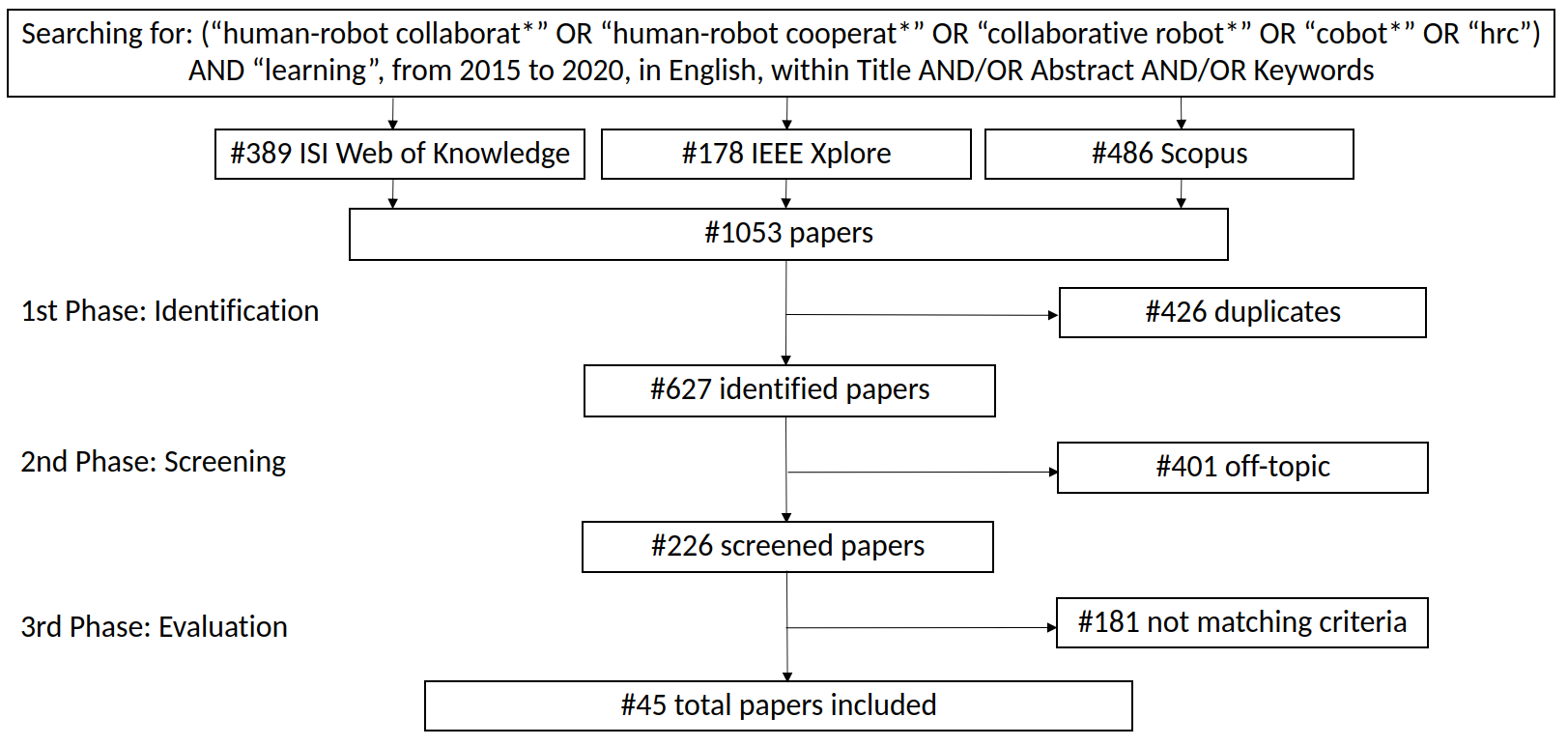}
  \hfill
  \caption{Process of paper selection performed in this review.}
  \label{workflow}
\end{figure}

Regarding point 3, an important exception involves collaborative assembly works (see section 3). Works of this kind were included, although the exchange of mechanical power is minimal. However, most of times this was due to the experiment being purposely approximated in this aspect. Indeed, these works focus more on the cognitive aspect of the interaction between a human and a robot, with no physical interface present, to properly look at the cognitive aspect of the problems. But, as these experimental validations can include physical interaction, it was decided to include the related papers, too.
The application of these selection criteria allowed us to achieve a final set of 45 papers (3 in 2015, 3 in 2016, 4 in 2017, 12 in 2018, 7 in 2019 and 16 in 2020), whose main elements are reported in Table \ref{table-wrap-39aebfabd1354e859fd12b3e49d75724}. The whole research and selection methodologies are summarised in Figure \ref{workflow} and all the papers, along with all the extracted aspects, are reported in Table \ref{table-wrap-39aebfabd1354e859fd12b3e49d75724}.

Figures \ref{tasks_hist}, \ref{metrics_hist}, \ref{all_cognitive}, \ref{all_ml} and \ref{sensing} show the trends of the features of the paper reported in this review. In these, the total number of instances are different from the papers' because some works show more than one instance of the same feature (see Table \ref{table-wrap-39aebfabd1354e859fd12b3e49d75724}).  

\begin{landscape}
\begin{table*}[!htbp]
\renewcommand{\arraystretch}{1.5}
\tiny
\caption{{Features extracted from the final set of papers. The presence of a ``+'' in an entry means that the elements are part of a unique framework, while a ``,'' denotes multiple ones. For the ``ML method'' column, the legend is: ANN=Artificial Neural Network, CNN=Convolutional Neural Network, Fuzzy=Fuzzy systems, DQN=Deep Q Network, GMM=Gaussian Mixture Model, HMM=Hidden Markov Model, MLP=Multi-Layer Perceptron, DNS=Dynamic Neural System, RL=Reinforcement Learning, RNN=Recurrent Neural Network, LSTM=Long Short Term Memory, VAE=Variational Autoencoder. Besides, for the same column, the numbers refer to which category they belong: A=Reinforcement learning, B=Supervised learning, C=Unsupervised learning. For the ``Cognitive ability'' column, the legend is: 1=Human effort, 2=Human intention, 3=Human motion, 4=Object of interest, 5=Robot actuation, 6=Robot decision making, 7=Task recognition.  For the ``Result'' column, the legend is: R1=Accuracy of movement, R2=Robustness, R3=Proof of concept, R4=High-level performance increase, R5=Reduction of human workload. For the ``Robotic platform'' column, the number in brackets highlight the number of degrees of freedom of the robotic platform used.} }
\label{table-wrap-39aebfabd1354e859fd12b3e49d75724}
\def\arraystretch{1}
\ignorespaces 
\centering 
\begin{tabulary}{\linewidth}{LLLLLLLLL}
\tbltoprule\\ 
\textbf{Authors} &
  \textbf{ML technique} &
  \textbf{Cognitive ability} &
  \textbf{Sensing} &
  \textbf{Interaction task} &
  \textbf{Robotic platform} &
  \textbf{Human role} &
  \textbf{Robot role} &
  \textbf{Key result}\\
\unskip~\cite{Ahmad2020} &
   CNN (B) &
   Position of object of interest (4) &
   Vision &
   Object handling &
   Custom &
   The human positions the tool &
   The robot actuates itself changing orientation of flexible tip &
   System was 60\% accurate in adjusting properly according to the object of interest (R1)\\
\unskip~\cite{Akkaladevi20191491} &
   Interactive RL (A)&
   Decision making (6) &
   Vision &
   Collaborative assembly &
   Universal Robots UR 10 (6) &
   The user has its own part in the process, but can also actively alter the decisional process of the robot &
   The robot observes the scene, understands the assembly process and chooses action to perform &
   The system is able to handle deviations during action execution (R2)\\
\unskip~\cite{ChenM2020} &
   SARSOP (A)&
   Human trust (2) &
   Vision &
   Object handling &
   Kinova Jaco (6) &
   The human either stays put or prevents the robot from executing an action, if not trusting it &
   The robot clears tables from objects &
   The RL model matches human trust to robot's manipulation capabilities (R3)\\

\unskip~\cite{ChenX2020stiffness} &
   CNN (B)&
   Human intention (2) &
   Muscular activity &
   Object handling &
   Rethink Robotics Baxter (7) &
   The human jointly lifts an object with the robot and moves it to a different position &
   The robot detects the human intention and follows its movement, keeping the object balanced  &
   The robot is successful in the task execution (R3)\\
\unskip~\cite{ChenX2020Neural} &
   ANN (B)&
   Human motion (3) &
   Accelerometry &
   Collaborative manufacturing &
   Rethink Robotics Baxter (7) &
   The human holds one end of the saw and saws different objects in different ways &
   The robot holds the other end of the saw and tracks its movement by recognising parameters &
   The devised system shows smoother performance than the comparison approach (R4)\\
\unskip~\cite{Chi2018} &
   GMM (C)&
   Position of object of interest (4) &
   Accelerometry &
   Object handling &
   Custom &
   The human positions the end effector of the robot with a certain pose and holds the robot in place &
   The robot detects its own end effector and adjusts it with its distal DoFs &
   Robot performs positioning with lower contact forces on target than the ones performed by human (R5)\\
\unskip~\cite{Choi2018} &
   Density-matching RL (A)&
   Decision making (6) &
   Vision &
   Object handover &
   Robotis Manipu- lator-X (7) &
   The human goes to grab the object held by the robot, but decides to change way to grab it at a certain point in the interaction &
   The robot recognizes change in user's motions and determines its actions accordingly &
   The robot recognizes change in user's motions and changes its actions accordingly (R3)\\
\unskip~\cite{Chung2015} &
   RL (A)+ Bayes (B) &
   Decision making (6) &
   Vision &
   Collaborative assembly &
   Gambit (6) &
   The human gazes at an object and goes to fetch it &
   The robot understands the intention and goes to reach the same object jointly with the human; if it is not confident about the action to perform, it can request help to the human &
   The interaction with the human improved the robot performance (R3)\\
\unskip~\cite{Cunha2020} &
   DNS (B)&
   Task recognition (7) &
   Vision &
   Collaborative assembly &
   Rethink Robotics Sawyer (7) &
   The human builds a system of coloured pipes jointly with the robot &
   The robot understands the current step of the assembly sequence and performs a complimentary action; it also gives suggestions to the user about next steps it should perform &
   The model is able to adapt to unexpected scenarios (R2)\\
\unskip~\cite{Deng2017} &
   Q-learning (A) &
   Decision making (6) &
   Vision &
   Object handling &
   Universal Robots UR5 (6) &
   The human jointly lifts an object with the robot and moves it repeatedly from point A to point B &
   The robot detects the human intention and follows its movement, keeping the object with a certain orientation &
   After some episodes, the robot is able to learn the path performed by the human and follow it properly (R3)\\
\unskip~\cite{Ghadirzadeh2016} &
   Q-learning (A) &
   Decision making (6) &
   Vision &
   Object handling &
   Willow Garage PR2 (7) &
   The human holds one end of a plank with a ball on top of it and moves it &
   The robot holds the other end of the plank and moves jointly with the human to prevent the ball from falling  &
   The measured force of interactions are lowered thanks to the learning process (R5)\\

\tblbottomrule 
\end{tabulary}\par 
\end{table*}

\end{landscape}

\begin{landscape}
\begin{table*}[!htbp]
\tiny

\def\arraystretch{1}
\ignorespaces 
\centering 
\begin{tabulary}{\linewidth}{LLLLLLLLL}
\tbltoprule \\
\textbf{Authors} &
  \textbf{ML technique} &
  \textbf{Cognitive ability} &
  \textbf{Sensing} &
  \textbf{Interaction task} &
  \textbf{Robotic platform} &
  \textbf{Human role} &
  \textbf{Robot role} &
  \textbf{Key result}\\

\unskip~\cite{Ghadirzadeh2020} &
   VAE (C)+ LSTM (B)+ DQN (A) &
   Decision making (6) &
   Vision &
   Collaborative assembly &
   ABB YuMi (7) &
   The human packages an object &
   The robot understands the actions performed by the human and plan its choices to assist in the assembly &
   The algorithm outperforms supervised classifiers (R4)\\
\unskip~\cite{Grigore2018} &
   HMM (B) &
   Task recognition (7) &
   Vision &
   Collaborative assembly &
   Rethink Robotics Baxter (7) &
   The human assembles a chair &
   The robot predicts the next step of the assembly and assists the human accordingly &
   Robot prediction capability matches human one (R3)\\
\unskip~\cite{Huang2018} &
   GMM (C)+ PI$^2$ (A) &
   Robot motion (5) &
   Vision &
   Object handover &
   IIT COMAN (4) &
   The human hands over an object to the robot under different conditions &
   The robot manipulator moves to reach the object &
   The robot performs the task respecting every constraint applied to it (R4)\\
\unskip~\cite{Kuang2018} &
   ANN (B) &
   Control variables (5) &
   Accelerometry &
   Object handling &
   Custom &
   The human grasps a tool held by the robot and moves it to a certain position &
   The robot adapts its movements to the trajectory imposed by the human &
   The system performs with higher safety, compliance and precision (R1)\\
\unskip~\cite{Lorenzini2018} &
   ANN (B)&
   Ground reaction force + centre of pressure (1) &
   Accelerometry &
   Object handling &
   KUKA LWR (7)  &
   The human lifts an object by one end &
  The robot grabs the object by the other end and helps the human lifting&
  The joint force exerted on the human were noticeably reduced (R5)\\
\unskip~\cite{Lu20201116} &
   LSTM (B)+ Q-learning (A) &
   Human intention (2), control variables (5) &
   Accelerometry &
   Object handling &
   FRANKA EMIKA Panda (7) &
   The human drags the object held by the robot &
   The robot detects the intention of the human and follows the trajectory forced by the dragging of the object &
   Interaction forces are noticeably reduced (R5)\\
\unskip~\cite{Luo2018} &
   GMM (C)&
   Human motion (3) &
   Vision &
   Collaborative assembly &
   Willow Garage PR2 (7) &
   The human touches pads of certain colour &
   The robot has to touch different pads than the human, avoiding undesired collisions with it &
   94.6\% of the trials were successful (R3)\\
\unskip~\cite{MaedaNeumann2017} &
   GMM (C)&
   Human motion (3) &
   Vision &
   Object handover &
   KUKA LWR (7) &
   The human passes an object to the robot, from one end of the table &
   The robot, from the opposite end of the table, recognizes what type of object is being passed, and receives it accordingly &
   The system demonstrated high accuracy in recognising the tool from human observation (R4)\\
\unskip~\cite{MaedaEwerton2017} &
   Interaction Probabilistic Movement Primitives (C)&
   Human motion (3) &
   Vision &
   Object handover &
   KUKA LWR (7) &
   The human is located at an end of a table and holds its hand to receive an object &
   The robot on human's adjacent side picks an object from human's opposite end of the table and hands the object to the human &
   Robot's positioning error lower than the one registered with standardized methods (R1)\\
\unskip~\cite{Munzer2018} &
   Relational Action Process (A)&
   Human intention (3), robot decisions (6) &
   Vision &
   Collaborative assembly &
   Rethink Robotics Baxter (7) &
   The human assembles a box &
   The robot classifies the user and its own actions and assists accordingly &
   Decision risk lowered thanks to devised approach (R2)\\
\unskip~\cite{MurataLi2018} &
   RNN (B)&
   Goal target (7) &
   Vision &
   Collaborative assembly &
   Softbank Robotics Nao &
   The human hits some objects with a known sequence &
   The robot recognizes the sequence and replicates it; the last part of the sequence must be done collaboratively &
   The robot was able to fulfill the task under the setup conditions (R3)\\
\unskip~\cite{MurataMasuda2019} &
   VAE (C)+ LSTM (B) &
   Goal image (7) &
   Vision &
   Object handover &
   Tokyo Robotics Torobo (6) &
   The human needs to move objects from a place to another &
   The robot looks at the human hand, understands the desired object and hands it over the human &
   The proposed framework allowed the robot to infer the human goal state and adapt to situational changes (R2)\\
\unskip~\cite{Nikolaidis2015} &
   Inverse RL (A)&
   Human type of worker (2) &
   Vision &
   Collaborative manufacturing &
   ABB IRB (6) &
   The human has to polish two sides of an object held by the robot &
   The robot recognizes the type of user and adjusts itself according to the information &
   The system proves to be more efficient than a person telling the robot what actions to perform (R3)\\
\unskip~\cite{Peternel2019} &
   Gaussian Process Regression (B)&
   Human muscular activity (1) &
   Vision+ Accelerometry &
   Collaborative manufacturing &
   KUKA LWR (7) &
   The human has to polish or drill surfaces held by the robot &
   The robot orients the surfaces to maximize certain parameters of interest &
   The system was able to reduce muscle fatigue of human (R5)\\

\unskip~\cite{Roveda2019} &
   ANN (B)&
   Robot level of assistance (6) &
   Vision+ muscular activity+ Accelerometry&
   Object handling &
   KUKA LWR iiwa 14 820 (7) &
   The human has to lift an object &
   The robot assists the lifting of an object, recognizing the type of assistance required &
   The overall muscular activity of the human is reduced (R5)\\

\tblbottomrule 
\end{tabulary}\par 
\end{table*}

\end{landscape}

\begin{landscape}
\begin{table*}[!htbp]
\tiny

\def\arraystretch{1}
\ignorespaces 
\centering 
\begin{tabulary}{\linewidth}{LLLLLLLLL}
\tbltoprule \\
\textbf{Authors} &
  \textbf{ML technique} &
  \textbf{Cognitive ability} &
  \textbf{Sensing} &
  \textbf{Interaction task} &
  \textbf{Robotic platform} &
  \textbf{Human role} &
  \textbf{Robot role} &
  \textbf{Key result}\\
  \unskip~\cite{Roveda2020} &
   Model-based RL (A)+ MLP (B) &
   Control variables (5) &
   Accelerometry &
   Object handling &
   KUKA LWR iiwa 14 820 (7) &
   The human has to lift an object &
   The robot assists the lifting of an object &
   The robot fullfills its task, while respecting embedded safety rules (R3)\\
\unskip~\cite{Rozo2015} &
   GMM (C) &
   Task recognition (7) &
   Accelerometry &
   Object handling &
   Barrett Technology WAM (7) &
   The human grabs an object already held by the robot  &
   After offline demonstrations, the robot moves together with the human's hand to the target position &
   Robot shows higher compliance thanks to the devised solution (R4)\\
\unskip~\cite{Rozo2016a} &
   HMM (C)&
   Robot motion (5) &
   Accelerometry &
   Object handover, object handling &
   Barrett Technology WAM (7) &
   The human holds an object of interest &
   After offline demonstrations, the robot pours the liquid from a bottle it is already holding &
   The system improves reactive and pro-active capabilities of the robot (R4)\\
\unskip~\cite{Sasagawa2020} &
   LSTM (B)&
   Control variables (5) &
   Accelerometry &
   Object handling &
   Haptic Device Touch (3) &
   The human holds a fork and uses it to collect food &
   The robot holds a spoon and helps the human scooping the same piece of food &
   The robot is successful in the task execution (R3)\\
\unskip~\cite{Shukla2018} &
   Proactive Incremental Learning (B)&
   Decision making (6) &
   Vision &
   Object handover &
   KUKA LWR (7) &
   The human gestures to the robot for the action and gazes at the object of interest &
   The robot recognizes the command, grabs the required object and hands it over the human &
   The devised system has minimized the robot attention demand (R4)\\
\unskip~\cite{Tabrez2019} &
   RL (A) &
   Decision making (6) &
   Vision &
   Collaborative assembly &
   Rethink Robotics Sawyer (7) &
   The human moves pieces on a surface according to certain rules &
   The robot moves other pieces according to human's actions following a shared objective, giving reasons for the actions performed  &
   The users found the robot more engaging because of the motivations given (R3)\\
\unskip~\cite{Tsiakas2017423} &
   Interactive RL (A) &
   Decision making (6) &
   Vision &
   Collaborative assembly &
   Barrett Technology WAM (7) &
   The human has to move blocks from a place to another &
   The robot jointly moves other pieces at the same time &
   The robot was successful in the task (R3)\\
\unskip~\cite{Unhelkar2020} &
   RL (A) &
   Decision making (6) &
   Vision &
   Collaborative assembly &
   Universal Robots UR10 (7) &
   The human has to follow different steps to prepare a meal &
   The robot has to act concurrently with the human in the preparation; it communicates the steps it is about to perform to the human &
   The required time for task completion is lower during human-robot collaboration (R4)\\
\unskip~\cite{Spaa9197296} &
   kNN (B)&
   Human hand pose (3) &
   Accelerometry &
   Object handling &
   KUKA LWR+ Custom (32) &
   The human has to support an object, while applying a rotation to it &
   The robot, by looking at the human hand pose, has to provide the required rotation &
   The devised system lowered biomechanic measurement (R5)\\
\unskip~\cite{Vinanzi2020} &
   Clustering (C)+ Bayes (B) &
   Human intention (2) &
   Vision &
   Collaborative assembly &
   Rethink Robotics Sawyer (7) &
   The human has to align 4 blocks, with a rule unknown to the robot &
   The robot recognizes human intention and grabs the remaining needed blocks &
   The robot achieved 80\% average accuracy in classifying the sequences being performed (R4)\\
\unskip~\cite{Vogt2016} &
   GMM (C)+ HMM (C) &
   Human motion (3) &
   Vision &
   Collaborative assembly &
   Universal Robots UR5 (6) &
   The human has to assemble a tower with blocks &
   The robot understands human's motion and acts accordingly to assist during the task &
   The robot is successful in assisting the human (R3)\\
\unskip~\cite{WangDiekel2018} &
   Inverse RL (A) &
   Robot actions (6) &
   Accelerometry+ muscular activity+ speech&
   Collaborative assembly &
   Staubli TX2-60 (6) &
   The human has to put concentric objects on top of each other  &
   The robot receives the inputs and assists the human in the process, by grasping the proper object; there is no specific order for the assembly steps execution &
   The robot matches most of times the behaviour expected by the human (R3)\\

\tblbottomrule 
\end{tabulary}\par 
\end{table*}

\end{landscape}

\begin{landscape}
\begin{table*}[!htbp]
\tiny

\def\arraystretch{1}
\ignorespaces 
\centering 
\begin{tabulary}{\linewidth}{LLLLLLLLL}
\tbltoprule \\
\textbf{Authors} &
  \textbf{ML technique} &
  \textbf{Cognitive ability} &
  \textbf{Sensing} &
  \textbf{Interaction task} &
  \textbf{Robotic platform} &
  \textbf{Human role} &
  \textbf{Robot role} &
  \textbf{Key result}\\

\unskip~\cite{Wang2018279} &
   Extreme learning machine (B) &
   Human intention (2) &
   Accelerometry+ muscular activity+ speech&
   Object handover &
   Staubli TX2-60 (6) &
   The human is handing over an object to the robot  &
   The robot understands the intent of the human and acts accordingly &
   The approach resulted to be higher in accuracy than past works (R4)\\
\unskip~\cite{Wojtak} &
   DNS (B) &
   Decision making (6) &
   Vision &
   Object handover &
   Rethink Robotics Sawyer (7) &
   The human is assembling a system of pipes &
   The robot looks at the human, understands the step of the process it is currently at and passes the pipe needed by the human &
   The devised system improves its efficiency with respect to a standard approach (R4)\\
\unskip~\cite{Wu2020Adaptive} &
   Q-learning (A) &
   Control variables (5) &
   Accelerometry &
   Object handling &
   FRANKA EMIKA Panda (7) &
   The human moves the object held by the robot along a 2D surface &
   The robot keeps holding the object and follows the trajectory &
   The robot performs better in terms of precision of positioning and time efficiency (R1)\\
\unskip~\cite{Wu2020Shared} &
   Q-learning (A) &
   Control variables (5) &
   Accelerometry &
   Object handling &
   FRANKA EMIKA Panda (7) &
   The human moves an object held by the robot from a point to another repeatedly; in case of variation, it moves it to a third point &
   The robot follows the trajectory forced by the human on the object; if it goes to the third point, it plans another trajectory to reach the second point &
   The robot is successful in the task, as long as the human does not exert high forces on the object (R3)\\
\unskip~\cite{Yan20191390} &
   LSTM (B) &
   Human intention (2) &
   Vision &
   Collaborative assembly &
   Universal Robots UR5 (6) &
   The human is assembling different pieces together &
   The robot has to predict the intention of the human and assist in it accordingly &
   The robot achieves the task with lower error in positioning (R1)\\
\unskip~\cite{Zhang2020}&
   RNN (B) &
   Human pose (3) &
   Vision &
   Object handover &
   Universal Robots UR5 (6) &
   The human is working on the assembly of an object &
   The robot looks at the human, predicts its next pose and, according to that, moves pre-emptively to pick up a screwdriver and pass it to the human  &
   The robot is successful in the task execution (R3)\\
\unskip~\cite{Zhou2019} &
   DNS (C) &
   Human intention (2) &
   Vision+ muscular activity+ brain activity &
   Object handover &
   Barrett technology WAM (7) &
   The human is assembling a chair &
   The robot, positioned at the side of the human, predicts the human's intention and hands over the related tool &
   Optimal settings for best system performance were identified (R4)
  \\
\tblbottomrule 
\end{tabulary}\par 
\end{table*}
\end{landscape}

\section{Human-robot collaboration tasks}

This review aims to provide insights about the use of machine learning in human-robot collaboration studies. Therefore, the interaction tasks investigated in the selected papers, together with the main research validations that were pursued, are reported and analysed in this section. All these works make use of a robotic manipulator as cobot and envisions the presence of a single user in the experimental validation. If the robotic architecture is a more complex robot, only its manipulation capabilities were exploited \unskip~\cite{Ghadirzadeh2016,Huang2018,Luo2018,MurataMasuda2019}. According to the features they share and the level of focus on the cognitive rather than the physical side of interaction, these works were subdivided in different categories (see Figure \ref{tasks_hist}).

\label{experiments}

\subsection{Collaborative tasks}
\label{tasks}
\begin{figure}[H]
  \centering
  \includegraphics[width=\textwidth]{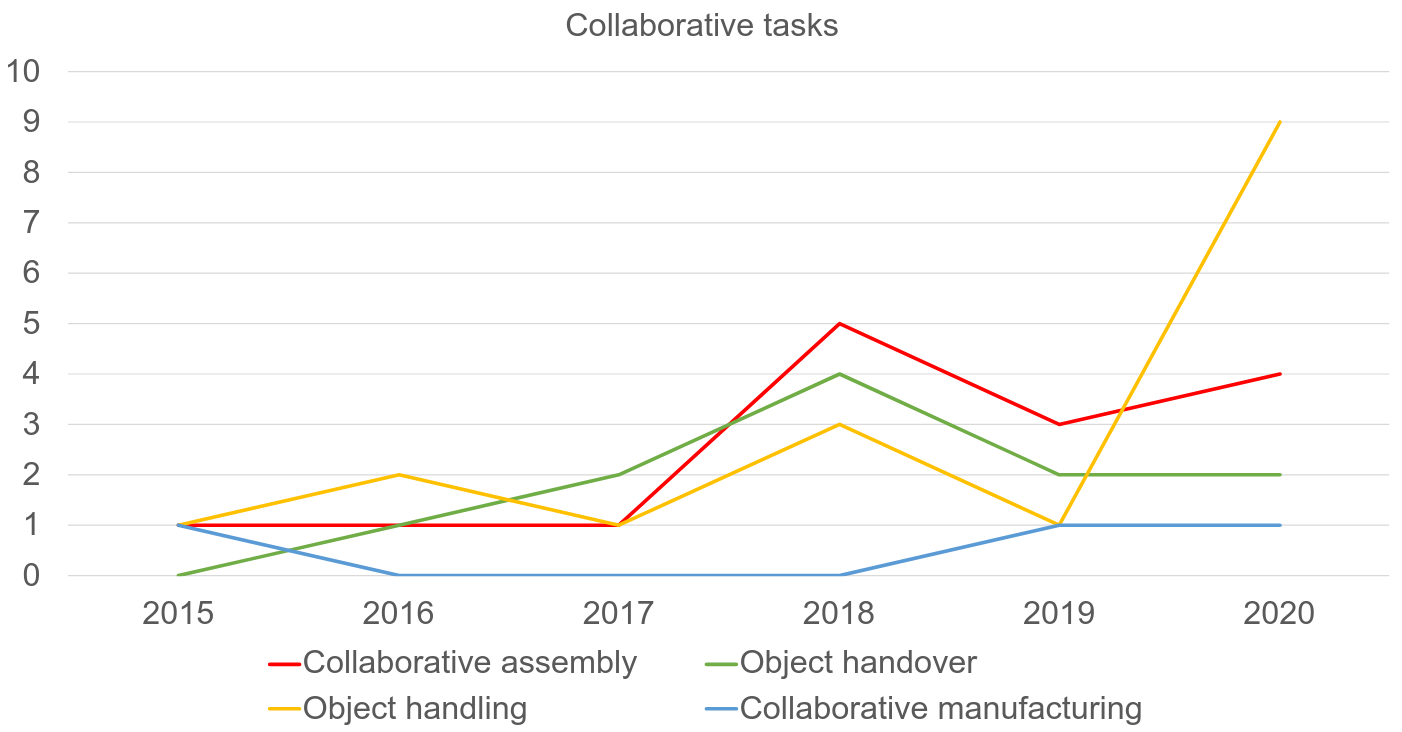}

  \hfill
  \caption{Time trends of the collaborative tasks explored in HRC (15 collaborative assemblies, 11 object handovers, 17 object handling and 3 collaborative manufacturing works).}
\label{tasks_hist}
\end{figure}
\label{collaborative}

Some works focus on the exploration of the cognitive side of the interaction. One of the most prominent types of cognitive interaction in HRC is the \textit{collaborative assembly}. In this category, the user has to assemble a more complex object by means of sequential sub-processes. The robot has to assist the user in the completion of them. This matches the definition of collaboration used before (see Section \ref{introduction}): human and robot work on the same task, in the same workspace, at the same time. In the majority of these works, the robot recognizes the step of the sequence the user is currently working at and performs an action to assist the user in the same sub-tasks. For example, in Cuhna et al. \unskip~\cite{Cunha2020}, the user is assembling a system of pipes; the robot has to understand in which step the user is currently at and grab the pipe related to that step of the assembly. Munzer et al. \unskip~\cite{Munzer2018} have the robot help the user assembling a box, through the understanding of what it is doing. In other cases, instead, the robot predicts the next step of the assembly process and prepare the scenario for the user to perform it later. In Grigore et al. \unskip~\cite{Grigore2018}, the robot predicts the trajectory the human will perform while seeking assistance from the robot, and executes an action accordingly. In other cases, the robot can also perform an action on the object of interest to conclude the process started by the user. In Vinanzi et al. \unskip~\cite{Vinanzi2020}, the user is building a shape by means of blocks and the robot, thanks to the information it gathers, finishes the construction with the remaining blocks.

Another categorization in which the cognitive aspect of the interaction still carries most of the focus is \textit{object handover}. Here, works focus on a specific part of the interaction, without considering a more complex task in the experimental design. In these experiments, user and robot exchange an item. Most of times, the robot hands it over to the user. It is important for the robot, in this case, to understand the user's intentions and expectations in the reception of the object. In Shukla et al. \unskip~\cite{Shukla2018}, the robot needs to understand the object desired by the users among the available ones and hand it over. In different terms, Choi et al. \unskip~\cite{Choi2018} design the robot to adjust to a change in the intentions of the user, while receiving the object. Conversely, there are cases in which the user hands the object over to the robot \unskip~\cite{Huang2018, MaedaEwerton2017}. 


Another family of works delves more into the physical side of the human-robot collaboration. In these, information collected from the environment is used to extract simpler high-level information than the previous cases. Besides, the same information is also used to control the actuation of the robot in the physical interaction. In this type of tasks, the exchange of mechanical power with the user is of the utmost importance. Indeed, in these cases, the robot can exert higher forces, according to the task. So, further aspects related to the safety of the interaction come into play.

The most complex interaction task of this family is the \textit{collaborative manufacturing}. Here, the robot applies permanent physical alterations to an object in collaboration with the human, as part of a manufacturing process. In this case, the exchange of interaction forces between human and robot are crucial. In Nikolaidis et al. \unskip~\cite{Nikolaidis2015} and Peternel et al. \unskip~\cite{Peternel2019}, the robot has a supportive role: it holds the object and orients it, according respectively to the intentions of the user and the parameters of interest to optimize, to assist the user in the polishing task being performed (\unskip~\cite{Peternel2019} considers drilling, too). However, in Chen et al. \unskip~\cite{ChenX2020stiffness}, the robot takes a more active role, applying forces jointly with the user, to hold a saw from one of its ends to cut an object.

In \textit{object handling}, the most recurrent interaction in this family, the task starts with the robot and the user having already jointly grabbed an object from different sides. Then, they move it together to another point in space, without any of them letting go of it. Of course, it is possible to gather high-level understanding from the collected information. Roveda et al. \unskip~\cite{Roveda2019} implemented a robotic system understanding which type of assistance is required by the user in the interaction and performs it accordingly. Similarly, in Sasagawa et al. \unskip~\cite{Sasagawa2020}, a robot holds a spoon and, according to what the user does with a fork, they scoop different types of food together, in the related way. However, it is clear how, in this type of interaction, the physical component is predominant. In most cases, the robot simply follows the movements of the user in the displacement of the object \unskip~\cite{Kuang2018,Lu20201116,Spaa9197296}. In more complex versions of this task, the robot, while the object is being displaced, maintains certain constraints about it. For instance, in Ghadirzadeh et al. \unskip~\cite{Ghadirzadeh2016}, the robot needs to maintain a rolling object on top of another one it is holding in the collaboration. Likewise, Wu et al. \unskip~\cite{Wu2020Adaptive} have the object move only along a set surface. Two object handling works address the problem of fine positioning. In these cases, the user drags the robot's end effector inside a region of interest. The robot, then, according to the collected information, adjusts its own end-effector, while still being held by the user, to perform the exact position and orientation of it required by the task \unskip~\cite{Ahmad2020,Chi2018}. 

Looking at Figure \ref{tasks_hist}, the evolution through time of the amount of works in the depicted categories of collaborative tasks is reported. It is possible to appreciate that collaborative assembly experiments have a constantly increasing trend. This is evidence that robotic scientists are starting to consider more complex tasks, where the interplay between decisions of the user and the robot becomes crucial. Robots are starting to shift from being tools that have to follow user's movements to proper collaborative entities in the scene. Of course, works related to simpler interactions from a cognitive point of view, that is object handling and object handover, are not decreasing in their amount. Besides, a steep increase of object handling works from 2019 and 2020 was registered. Collaborative manufacturing works are still at early stages, being particularly specific categories of tasks, but their number is increasing, as well.
    
\subsection{Metrics of collaborative tasks}
\begin{figure}[H]
  \centering
  \includegraphics[width=\textwidth]{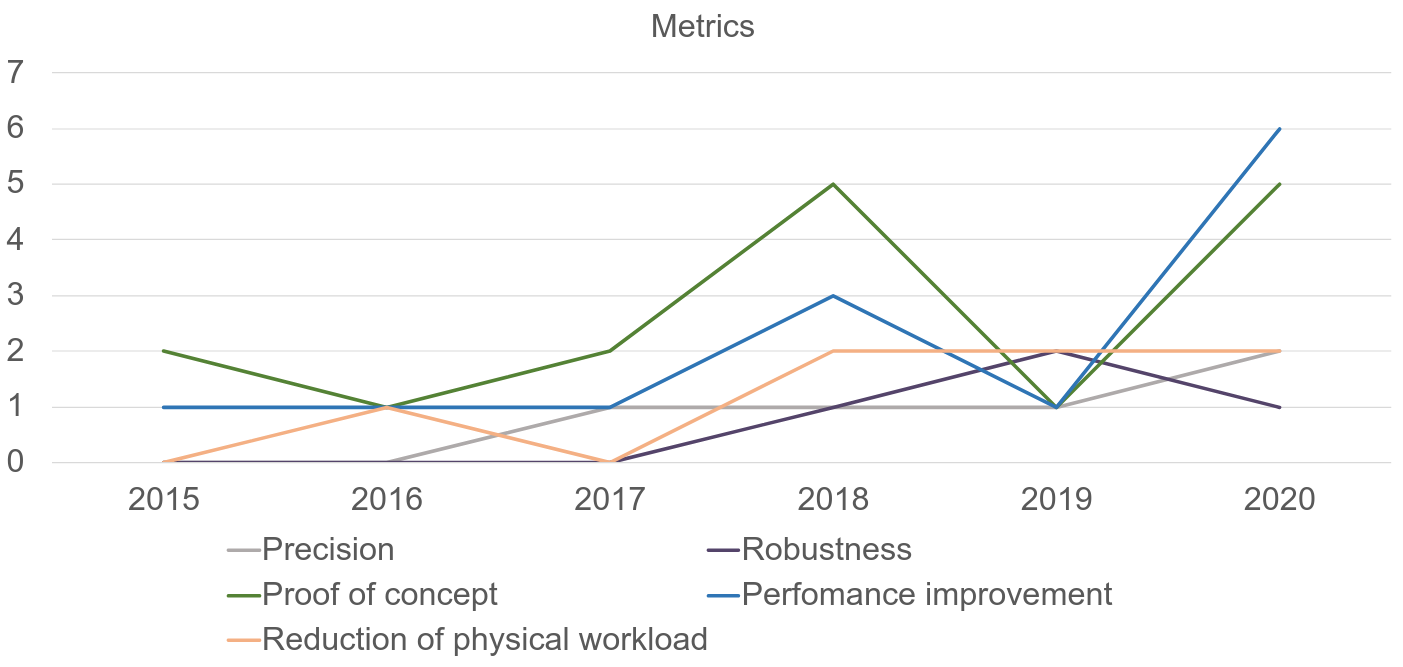}
  \hfill
  \caption{Time trends of the metrics employed in the experimental validations (5 related to precision, 4 to robustness, 17 to proof of concept, 13 to performance improvement and 7 to reduction of physical workload).}
\label{metrics_hist}
\end{figure}

In a research study, a goal is pursued through an experimental validation. According to the case, the result can vary in its nature. However, it is possible to come up with a categorization for them, as well (see Figure \ref{metrics_hist}).

The first reported set of results are referred to the precision of movement of the robot. Here, the good quality of the works was assessed by measuring the accuracy of the movement performed by the robot during the interaction. For the type of objectives being pursued in these papers, such quantity is very crucial. In most cases, they measure the positioning of the robot's end effector at the end of the movement, usually when it interacts with the object of interest or the user, like in object handling or object handover, respectively \unskip~\cite{Ahmad2020,MaedaNeumann2017}. In another case, as the study was related to object handling, the movement throughout the interaction was measured \unskip~\cite{Wu2020Adaptive}. 

In another set of works, robustness is the maIndeedor of interest. In general terms, this is the capability of robot to handle the collaboration in case something unpredictable occurs. This can happen when steps of a process or ways of interacting are expected, instead. In Cuhna et al. \unskip~\cite{Cunha2020}, one piece of the pipes needed for the assembly was purposely removed to evaluate how the robot could handle such unexpected scenario. However, it is also possible that the situation is built so that the robot has not any prior knowledge about the sequence of the steps to be executed \unskip~\cite{MurataMasuda2019}. In that case, every action performed by the user is an unexpected scenario the robot needs to address.

A third of the analysed works can be considered as a proof of concept of the designed system. In this sense, for the validation of the result, no quantitative measurements of the performance are provided. Indeed, main objective of these papers is to demonstrate that the devised systems can work in a way that is considered useful to the people who interact with the robot. Consequently, if they provide measurements about the outcome of the works, they are of a qualitative kind, about the perception users had about the robot improving the work performance. For example, Chen et al.\unskip~\cite{ChenM2020} conclude that most of times the user can put the required trust on the robot performing the task alongside it. In Luo et al. \unskip~\cite{Luo2018}, the robot met the expectations of the user about task completion. 

In more than a dozen of cases in the set of selected papers, a proof of better performances of the illustrated system is provided by means of comparison with a base case. The latter can be interpreted in multiple ways. For instance, improvement can be demonstrated by comparing the accuracy of the proposed ML system with the one achieved by a supervised classifier \unskip~\cite{Ghadirzadeh2020,Vinanzi2020}. Conversely, in Zhou et al. \unskip~\cite{Zhou2019}, different configurations of the same system were tested to highlight the one with best performances.

Finally, another subset of papers aimed at reducing the human physical workload during task completion. A common aspect between these papers is the use of biomechanical variables that were minimised using the robotic system, to validate the efficacy of the proposed solution \unskip~\cite{Lorenzini2018}. Another portion of them also looks at the reduction of the interaction forces between robot and user \unskip~\cite{Chi2018,Ghadirzadeh2016}. 

Regarding the metrics (see Figure \ref{metrics_hist}), it is highlighted that majority of these works still present qualitative results, as proof of concept of the devised cognitive system in HRC by means of machine learning techniques. However, it is interesting to witness that this kind of results was slightly surpassed by the ones given in terms of system performance improvement with respect to a given baseline. The other categories of result are more specific, so as a consequence, their number is lower. However, their number is increasing, but not to the same extent of works with proof of concept or performance improvement as result.

\section{Robot's cognitive capabilities}
\label{variables}
Depending on the interaction being addressed, every study focuses on the design and evaluation of a different cognitive or behavioural capability of the robot. This clearly is the most varied feature of the selected HRC papers. This is strictly related to the researchers’ interpretation of the problem and the solution they produce. In works in which the behavioural block is built through machine learning, an instance of such cognitive capabilities is represented by the outcome itself of these algorithms. In a low level of cognitive reasoning, these can be used as input to control the robot actuation. In more complex cognitive processing, they can be used to select ranges of control inputs for the robot.  

\begin{figure}[H]
  \centering
  \begin{subfigure}{0.50\textwidth}
    \includegraphics[width=\textwidth]{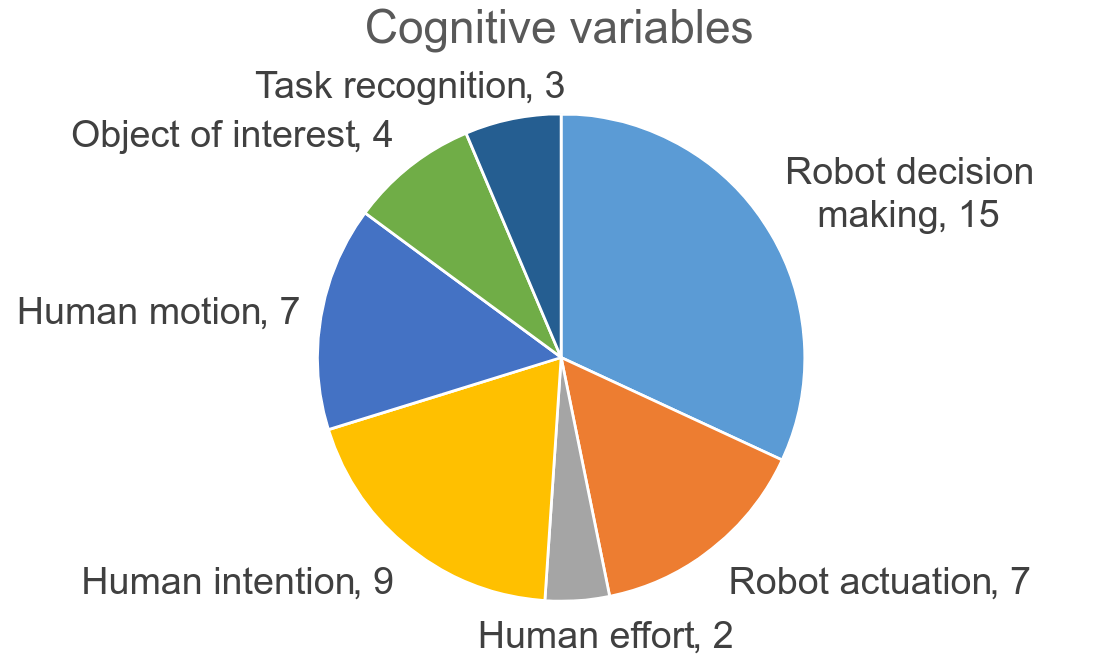}
    \caption{ }
    \label{pie_cognitive}
  \end{subfigure}
  \hfill
  \begin{subfigure}{0.49\textwidth}
  \centering
    \includegraphics[width=\textwidth]{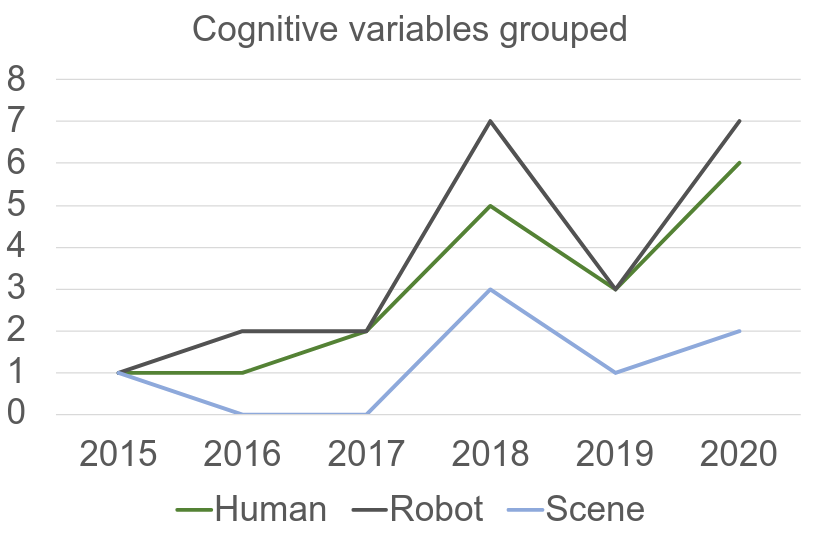}
    \caption{ }
    \label{cognitive_time}
  \end{subfigure}
  \caption{Results from the selected set of papers related to the modelled cognitive variables. In Subfigure \ref{pie_cognitive}, for each circular sector, the related category and its occurrence in the selected set of papers are shown. Subfigure \ref{cognitive_time} shows their time trends. In doing so, the first two categories, the three central ones and the last two were grouped together, according to them being related to aspects of the robot (22 works), the human user (18 works) or the scene (7 works).}
  \label{all_cognitive}
\end{figure}

An effort was made in this survey to group the investigated cognitive skills into categories, to highlight common aspects. In Figure \ref{all_cognitive}, it is possible to appreciate such a grouping. The specific cognitive variables are reported case by case in the third column of Table \ref{table-wrap-39aebfabd1354e859fd12b3e49d75724}, along with a number specifying the category they were assigned to.

Half of the cognitive variables are related to aspects of the robot itself. In most of these cases, they are represented by the robot decision about the next action to perform during the interaction. They are primarily chosen by considering the state of the interaction at the current time and assessing the return achieved by taking an action rather than another one. This statement inherently implies how well this way of modelling the interaction is prone to be tackled through reinforcement learning \unskip~\cite{Deng2017,Shukla2018, Tabrez2019} and graphical models \unskip~\cite{Chung2015,Unhelkar2020} (see Section \ref{ML}). In an alternative interpretation of this, ranges of interaction levels for the robot, instead of specific actions to perform, are considered \unskip~\cite{Roveda2020}.

Referring to a lower level of cognitive processing mentioned before, in other cases the machine learning algorithms produced the actual control variables to handle the actuation of the problem. Also in these cases, a modelling of the situation like the one described above is used \unskip~\cite{Wu2020Adaptive,Wu2020Shared}. In an alternative interpretation, Huang et al.\unskip~\cite{Huang2018} and Rozo et al. \unskip~\cite{Rozo2016a} designed an intermediate step before the output produced by the machine learning algorithms. Indeed, they used machine learning to model the motion of the robot itself. Based on that, they used their controller to produce the control variables to handle the subsequent steps of the human-robot collaboration.

The other half of the papers model factors related to the other side of human-robot collaboration, that is the human one. The human-related capability with the highest level of information provided are the ones grouped under human intention. This denomination covers a wide range of cognitive capabilities that assume different shades of meaning, according to the case. In many of them, it is interpreted as the intention of the user towards the task ahead. In two works \unskip~\cite{Vinanzi2020,Yan20191390}, the robot understands the intention of the user about the blocks to grab for the assembly and, based on that, performs the assistive action with the remaining ones needed to complete the task. In a particularly interesting case, instead of the user's intention, the different styles they would apply in completing the same task are modelled \unskip~\cite{Nikolaidis2015}. In another case, instead of considering the attitude of the user towards the task, the one towards the robot performance itself is investigated: Chen et al. \unskip~\cite{ChenM2020} design the trust the human puts in the robot, while working together in a table clearing task. So, human trust towards robots clearly is a generally relevant topic not only in human-robot interaction, but also in its specific declination of collaboration.

A different school of thought in modelling human factors focuses on the motions performed by the user during the experiment. This is because they also use this information as an initial starting point for the range of motions available to the robot, which is then fine-tuned by the higher-level information provided by the machine learning algorithms. Most of them analyses human motion in all its aspects \unskip~\cite{ChenX2020Neural, MaedaNeumann2017}. Some of them especially focus on the human skeleton pose \unskip~\cite{Zhang2020} or the hand pose \unskip~\cite{Spaa9197296}.

A minority of the works related to the use of human factors focuses on the modelling of the human effort. In general terms, this is defined as the physical fatigue the user undergoes while performing the task during the interaction. These cognitive variables, mostly inherited from human biomechanics, are used as quantities to minimise by the control routine following the machine learning algorithm. In Lorenzini et al. \unskip~\cite{Lorenzini2018}, the ground reaction force exerted on the user, along with his/her centre of pressure, are used by the robot to find the optimal way of leveraging the effort during a lifting task. In Peternel et al. \unskip~\cite{Peternel2019}, through measurement of the muscular activity of the user, the human fatigue is modelled and used as parameter of interest to regulate the collaboration.

Interestingly, a small subset of works in literature controls the human-robot collaboration without modelling parameters from either the human or the robot, but from the scenario itself. However, they constitute a small amount in the plethora of works. This is to be expected: one of the main objectives of human-robot interaction is the understanding of the mutual perception between a robot and a human. So, this can be more effectively surveyed by modelling the two components of the interplay, rather than focusing on the task itself. However, it is possible to model the interaction in these terms and get valid results in the experimental validation, as well. Some of these works model the position of the object of interest of the collaborative task, as the position of the robot from the object was relevant to the task \unskip~\cite{Ahmad2020,Chi2018}. Others, instead, recognize from the scene the task that has to be performed by the robot \unskip~\cite{Cunha2020,Grigore2018,Rozo2015}. Two cases built an internal representation of the final goal itself, by looking at the scene \unskip~\cite{MurataMasuda2019,MurataLi2018}.

The time trends (see Subfigure \ref{cognitive_time}) confirm that scene-related variables are used less with respect to the ones modelling human and robot aspects. Furthermore, it is possible to notice a slight predominance of robot-related variables with respect to human ones.
    
\section{Machine learning techniques}

\label{ML}

\begin{figure}[H]
  \centering
   \includegraphics[width=\textwidth]{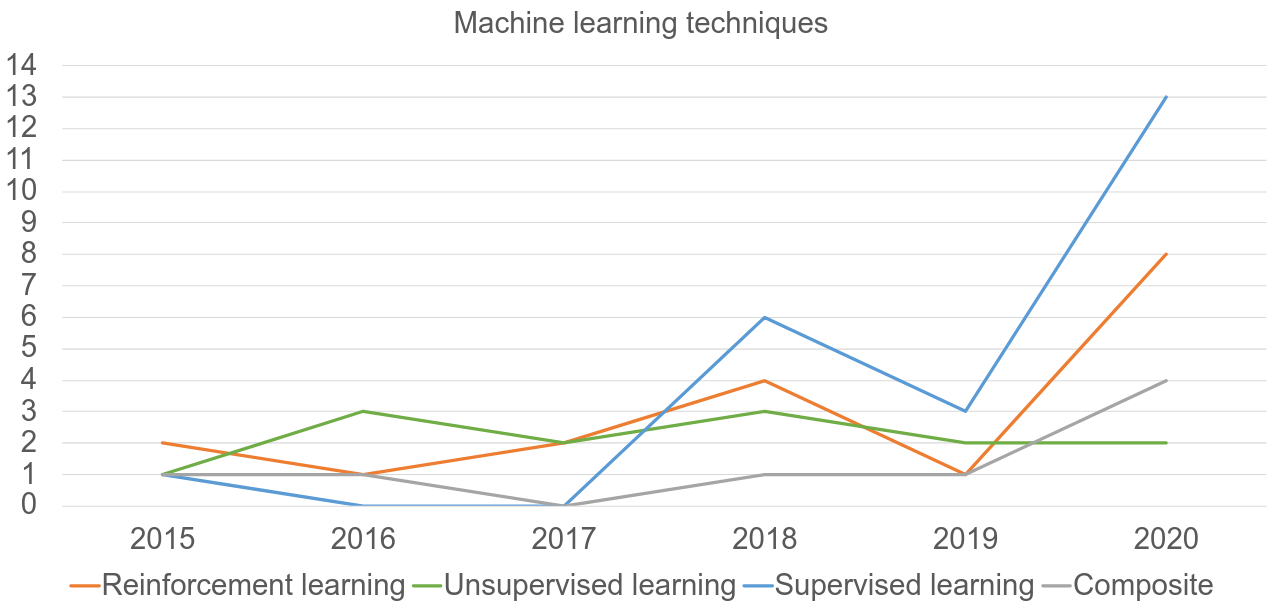}
  \hfill
  \caption{Results from the selected set of papers related to the machine learning algorithms (18 cases of reinforcement learning, 12 of unsupervised learning and 23 of supervised learning). Besides, there have been 1 case of deep learning model in 2018, 3 in 2019 and 8 in 2020. The "\textit{Composite}" series
 is related to the works that made use of more than one model as part of the same work (8 cases in total).}
  \label{all_ml}
\end{figure}

This section delves into the ML techniques used in the selected set of papers, to detect their most recurrent trends in HRC. First, it is important to mention that majority of papers uses only one ML model. Only a small amount of the papers designed the behavioural block of their robotic systems by cascading a combination of different ones (see Figure \ref{all_ml}). The main reason for this is that human-robot collaborations are extraordinarily complex to model. Therefore, researchers tend to focus on one aspect of the inference from the sensed environment and take the others for granted, thanks to the design of the experimental validation.

Following the most common  terminology to subdivide machine learning algorithms \unskip~\cite{jo2021machine}, ML techniques reported in this review were grouped into three categories: reinforcement learning, unsupervised learning and supervised reinforcement learning. In Figure \ref{all_ml}, it is possible to visualise the trends of these groupings over time. Up to 2017, there was no predominance among the categories. In 2018, the number of uses of ML techniques increased. In that year there was a steep increase of supervised learning cases. After a decrease in 2019, they kept increasing in 2020. 
Reinforcement learning followed a similar trend, but with less cases. Conversely, unsupervised learning algorithms have always been a stable trend, smaller in the amount of cases than the other two categories.

Besides, it is possible to appreciate that there have been few works that made use of a composite ML system. However, an increase was registered in 2020.

\subsection{Unsupervised learning}
\label{unsupervised}

HRC models the dynamics between a human and a robot to achieve a common goal. Due to their complexity, it is not very feasible to design the interaction by following deterministic schemes. Hence, many works make use of probability theory to devise a human-robot collaboration. In ML, probability is often modelled through  graphical models. Here, conditional dependencies among the random variables that model the problem are represented through nodes connected one another when a conditional dependence between the variables is assumed. ML models based on probability are often trained through unsupervised learning (UL) algorithms, that make use of unlabelled data in the learning process.


The most recurrent type of UL model is the Gaussian Mixture model (GMM). Here, a preset number of multi-variate Gaussian distributions is fitted into the space of the training dataset during the learning phase, process that is called Gaussian Mixture Regression (GMR). In the related graphical model, the observation is assumed conditionally dependent from the parameters that model the distributions and a latent variable vector that tells to which distribution the data point is likely to belong to. These latent variables are also conditionally dependent from the parameters of the distributions. This ML model is trained in an unsupervised fashion through the Expectation Maximization (EM) algorithm \unskip~\cite{Chi2018}.
Then, the ML model classifies a new data point according to the distribution most likely to belong to. 

A variation of GMM, the task parameterized GMM (TP-GMM), is particularly useful in the field of robotics \unskip~\cite{Huang2018,Rozo2015}. Indeed, it allows to perform GMR considering different frames from which the observations are taken. These are included in the model through task parameters. Such algorithm returns the mixture of Gaussian fitting according to the frame that returns the best performance in the considered task, which is useful for adaptive trajectories in robotic manipulators. 

Another ML model commonly trained with an UL algorithm is the Hidden Markov model (HMM). Here, dependence of a random variable from its value at previous timestamp is assumed, so they are able to take time dependence into account. The variables can only be observed through indirect observations (reason why they are called hidden variables), that correspond to a node connected only to the related timestamp being considered. In a HRC scenario, this can be seen as a robot trying to understand user's intentions by looking at its movements. HMMs are trained through a variation of the EM algorithm for HMMs, the Baum-Welch algorithm \unskip~\cite{Grigore2018}. Rozo et al. \unskip~\cite{Rozo2016a} et al. used it to perform a pouring task while the human user is holding a cup, while Vogt et al. \unskip~\cite{Vogt2016} employed a HMM to handle the collaborative assembly between a human and a robot.

The structure of a HMM allows to take time dependence into account during the learning phase of the algorithm. This factor is paramount when it comes to HRC. Indeed, most actions, in a general interaction, are influenced by time, or rather said, the sequentiality of the actions taken before the interaction at current time. Hence, it is crucial for a ML algorithm and model to be able to consider it.

A third ML model appeared in the selection is the variational autoencoder (VAE), from the family of \textit{deep learning} (DL) models. They are models composed by a neural network (for more technical details about a neural network, see Subsection \ref{supervised}) with two or more hidden layers, that are capable of mapping very complex nonlinear functions. DL models have become progressively more popular in HRC in the last years (see Figure \ref{all_ml}). In VAEs, parameters of a latent multivariate distribution are learnt as output of a DL model. Such output is then used to model the distribution, whose sample is used by a different deep neural network to model a distribution of the original training dataset. This model is trained through minimization of the so-called ELBO bound \unskip~\cite{MurataMasuda2019}. These models can also be used to perform classification, although the learning process makes use of no labels. In this case, this way of learning is addressed as self-supervised learning. Because they only learn alternate representations of the inputs, these models cannot take time dependence into account. Indeed, they are rather used as first stage of composite ML system \unskip~\cite{Ghadirzadeh2020,MurataMasuda2019}. This style of use can be found for GMM, as well \unskip~\cite{Huang2018,MaedaNeumann2017,Rozo2015,Vogt2016}, as neither of these two models is able to consider time dependence.




\subsection{Reinforcement learning}
\label{RL}

Markov decision processes (MDP) are another type of graphical model. Here, an agent interacts with an environment under a certain policy and gets a reward an observation about the state of it. This situation is very relatable to HRC applications: the robot is represented by the agent and the collaborative scenario by the environment. These processes can be totally observable \unskip~\cite{Ghadirzadeh2016}, partially observable \unskip~\cite{Ghadirzadeh2020} or with mixed observability \unskip~\cite{Nikolaidis2015}. Around the model of MDP, a wide plethora of learning algorithms is built, which constitutes the family of reinforcement learning (RL). Main objective, in this case, is the maximization of a cumulative function of the reward over time, that is called the value function. Consequently, this highlights the intrinsic ability of RL algorithm to consider time dependence.

Such result is achieved in these works in different manners. Model-free is the most used family of RL algorithms, simply because they do not require a model of the transition of the environment from a state to the next one, which is hard to design beforehand. Most frequent algorithm of this kind is Q-learning \unskip~\cite{Deng2017,Ghadirzadeh2016, Wu2020Adaptive, Wu2020Shared}. It is a model-free RL algorithm that updates the value function Q arbitrarily initialised. During the learning phase, its values are then updated with a percentage of the current reward and the highest value depending on the action and possible future states. Most of the works that use Q-learning sticks to the traditional formulation. However, others try variations of it. Lu et al. \unskip~\cite{Lu20201116}, they used Q-learning with eligibility traces and fuzzy logic for their object handling task. 

Of course, Q-learning is not the only model-free RL algorithm. Inverse reinforcement learning tackles a problem that is almost opposite to the one addressed by Q-learning. Here, the training phase does not make use of a reward function; instead, it is fed with a history of past concatenated actions and observations, with the idea of coming up with a reward function that motivates them. This approach to a HRC problem is spurred by the fact that most of times it is hard to start the training with a reward function, due to the complexity of the problem. Interestingly, Wang et al. \unskip~\cite{WangDiekel2018} used an inverse RL algorithm, integrating robotic learning from demonstration in the learning phase of the algorithm, when it processes the received history of observations and actions. 

With a different strategy, Nikolaidis et al. \unskip~\cite{Nikolaidis2015} used interactive RL, in which the agent learns both from observations from the environment and a secondary source, like a teacher or sensor feedback. A similar approach can be found in robotics research, in which people that are expert in a certain field, evaluate the performance of the robot in fulfilling a certain task.

There are also other works which do not consider model-free algorithms, although in a smaller amount. Roveda et al. \unskip~\cite{Roveda2020} use a model-based RL algorithm, by modelling the human-robot dynamics by means of a neural network. In \unskip~\cite{Huang2018}, a PI$^2$ algorithm is used. Here, the search for an optimal policy is performed by exploring noisy variations of the trajectories generated by the robotic manipulator and updating the task parameters according to the cumulative value of a cost function, that can be seen as the opposite of a reward function.

In composite ML systems, RL uses high-level information provided by other ML models, mostly neural networks trained in a supervised fashion \unskip~\cite{Lu20201116,Roveda2020} (see Subsection \ref{supervised}) . A special case of this is called \textit{deep reinforcement learning}. Here, the DL model outputs the value, given the current state, to map a complex state space into the value space. The DL model is trained jointly with the RL algorithm itself, thus becoming part of it. In \unskip~\cite{Ghadirzadeh2020}, they used a deep Q-network to handle the HRC scenario. Although deep RL is promising in the field of robotics, its use is still rare in HRC and therefore encouraged.

\subsection{Supervised learning}

\label{supervised}

RL determines a policy for an agent to maximize a cumulative reward through learning by interaction with the environment. Although this approach is really representative of a robot acting in the real world, many researchers still prefer to adopt supervised learning (SL) as a strategy to build a cognitive model. 

Here, main difference from RL is the request from the model to produce a label as output, given a certain input. Here, the training is performed with labelled instances of a dataset, while this is not the case with RL. Indeed, in RL the agent learns by interacting with the environment itself, through a trial-and-error process. In supervised learning, instead, the system does not receive a new input that is consequence of its output, because the output does not directly affect the environment, due to it simply being a label. Indeed, when SL is used, the output is then used as a high-level information in further processing to select the behaviour for the robot to perform. This ensemble is simpler to design than RL, because modelling the environment and the interaction with it is not needed. Besides, in terms of behavioural design, SL gives more freedom to the researcher to develop the actual behaviour of the robot. Conversely, having to work with hard labelling, supervised learning lacks the balance between exploration and exploitation that RL is more suitable to provide, instead, as it does not make use of previously set pairs of inputs and targets. Besides, supervised learning most of times cannot incorporate time dependence.

Models based on probability distributions can be trained in a supervised fashion. Most common example is Naive Bayes classification, in which supervised training through Bayes theorem is performed. Vinanzi et al. \unskip~\cite{Huang2018} made use of it to recognise the human intention during a collaborative assembly. Alternatively, Peternel et al. \unskip~\cite{Peternel2019} used Gaussian Process Regression to predict the value of the force during a collaborative manufacturing task. Besides, Grigore et al. \unskip~\cite{Grigore2018} trained a HMM, ML model typically trained with unsupervised learning algorithm (see section \ref{unsupervised}), in a supervised way, instead.    

However, majority of works involving SL tend to focus on deterministic models. A popular one is artificial neural networks (ANN). These models are composed by feedforward layers of neurons, or perceptrons, units that compute a function of a weighted sum of inputs. The connections between neurons are trained in a supervised way through the backpropagation algorithm \unskip~\cite{ChenX2020Neural}. In their simplest version, ANNs are composed by an input layer, a hidden layer and an output layer. Because of this, they are generally not good for taking into account time dependencies, as their memory is really dependent on the dimensionality of the input. However, it is interesting to notice a pattern in the use of ANNs for HRC. Indeed, in the related works, they are used to produce output variables strictly related to the actuation of the platform in the continuous time domain, from a control system point of view \unskip~\cite{Roveda2019}. They can, for instance, produce coefficients of a Lagrangian control system \unskip~\cite{ChenX2020Neural}. In Lorenzini et al. \unskip~\cite{Lorenzini2018}, instead, the output was the variable they tried to minimise in their work to improve the interaction. Furthermore, there is also a tendency to combine MLPs with fuzzy logic \unskip~\cite{Kuang2018,Lu20201116,Roveda2019}. A less standard way to train ANNs was performed by Wang et al. \unskip~\cite{Wang2018279}. They employed extreme learning machine, a training modality for ANNs that include neurons that do not need training. 

\textit{Deep learning} models are largely employed in SL for HRC, as well. They make up for almost half of the works related to SL in the selected set of papers. Indeed, the structure of these models allows better separation of the hyperspaces related to the different classes of a ML problem. A class of DL models capable of incorporating time dependence efficiently is the recurrent neural network (RNN). Here, the output is backpropagated in the network as part of the input of the next iteration of the training phase, learning algorithm that is called backpropagation through time. This allows to learn time dependence, as the output of the ML model dependent on the previous input is fed into the current one. Therefore, memory of the first input on is present, even after having processed an ideally infinite number of samples. In Zhang et al. \unskip~\cite{Zhang2020}, this is used to process sequences of motion frames of the user in the experimental validation. Specifically, they instantiated multiple cascaded RNNs, which is a typical way of using them. In another case, Murata et al. \unskip~\cite{MurataLi2018} implemented a time-scaled version of an RNN to account for slow and fast dynamics of the collaboration.

A particular category of RNNs is the long short term memory (LSTM) networks. They differ from standard RNNs for the presence of non-linear elements in their structure. They are well-known to be able to catch long term dependencies \unskip~\cite{Lu20201116}. Similarly to RNNs, the use of cascaded LSTMs is also reported \unskip~\cite{Yan20191390}. Furthermore, there is a tendency in cascading the LSTM with other ML models that are trained by means of UL or RL. In the first case, they receive a processed information from a model incapable of incorporating time dependencies \unskip~\cite{MurataMasuda2019}, while in the second one they extract the high-level information, like the user's intention \unskip~\cite{Lu20201116,Roveda2020}, then used to tune the agent's observation.

Furthermore, the use of convolutional neural networks (CNN) is also reported in HRC, but in lower amount. They are NNs specifically designed for the processing of raw images, or more generally inputs composed by measurements from multiple sensors. Indeed, an image is a clear example of this, as each one of them associated to a different pixel of an image. In this case, time dependence is more difficult to catch with CNNs. To do so, you need to concatenate multiple images as a unique input; however, the capability of incorporating the dependence is heavily constrained by the dimensionality of the input.  Ahmad et al. \unskip~\cite{Ahmad2020} use a CNN to retrieve the position of the object of interest, caught through a camera. In Chen et al. \unskip~\cite{ChenX2020stiffness}, the electromyography of the user, measured through multiple sensors, is processed by a CNN.

Finally, another subset of works implements the reasoning block of a robotic system by using a dynamic system that simulates a network of actual neurons, also called dynamic neural system (DNS). NNs take inspiration from the way a neuron works to build a ML model that does not truthfully follow its dynamics. But it is possible to build ML models that mimic the firing dynamics of actual neurons very faithfully and that can be used to perform machine learning. Of course, DNS are not ML models per se, so the way they are trained strictly depends on the specific work. However, being time-variant systems, they can incorporate time dependence. DNS are inherited from neuroscience, and the most used in HRC applications is the Amari equation \unskip~\cite{Cunha2020,Wojtak}.

\section{Robot's sensing modalities}

\label{sensing}
Human-robot collaboration tackles complex interaction problems between a person and a robot in a scenario where a task needs to be jointly completed. Due to these conditions, the environment where the robot is deployed is classified as unstructured. Here, every instance of the interaction differs from each other, by making impossible for this situation to be handled with standardized control methods where strong assumptions on the robot's workspace must be made. The robot needs to sense the environment to adjust to variations of the interaction, like shifts of the object of interest in the collaborative task or different positions of the user's hand, to consider when an object must be handed in.

\begin{figure}[!htbp]
  \centering
    \includegraphics[width=\textwidth]{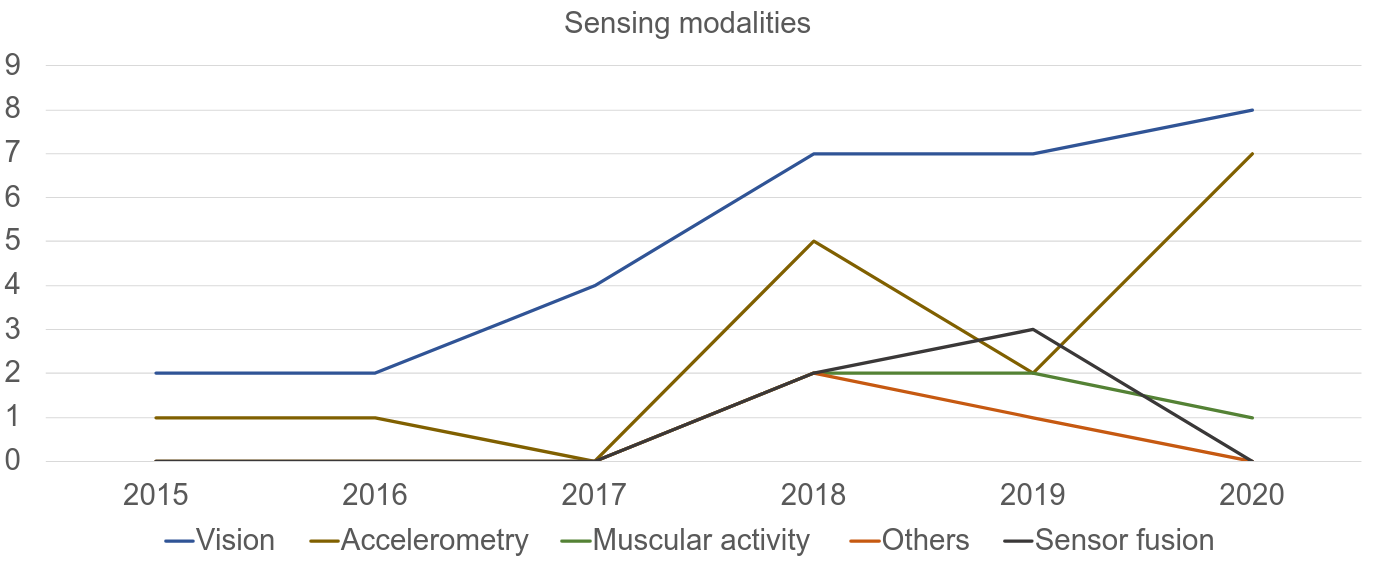}
    \label{time_sensing}
  \hfill
  \caption{Results from the selected set of papers related to the sensing modality (30 cases of use of vision, 16 of accelerometry, 5 of muscular activity and 3 of others). The "\textit{Others}" category is composed by one case of use of brain activity and two of speech. The series "\textit{Sensor fusion}" represents the trend of using multiple sensing in the same system, for a total of 5 cases.}
  \label{time_sensing}
\end{figure}

Figure \ref{time_sensing} shows that vision is the primary informative channel used in HRC. A main reason for this is that vision allows the system to potentially get information from different elements of the environment at the same time. For instance, Munzer et al. \unskip~\cite{Munzer2018} use vision to acquire information related to the robot and the users in the experiments. Indeed, through vision the robot becomes aware of the user's position in the environment and physical features. Being aware of its physical presence, it can compute position and orientation of the robot's end effector according to the specific user. So, the use of vision allows to cover many possible instances of the same interaction. 

The second most used sensing modality concerns accelerometry, from either the user \unskip~\cite{Wang2018279}, the scene \unskip~\cite{Rozo2015} or the robot itself \unskip~\cite{Wu2020Shared}. However, with a system capable of only measuring these quantities, stronger assumptions on the experimental setup are needed. For instance, in a case of object handling \unskip~\cite{Kuang2018} (see Subsection \ref{collaborative}), the robotic system will only be aware of the object of interest, and have no perception of its position in the environment. Therefore, during the experimental validation, collision avoidance with surrounding environment will have to be ensured through pre-setting of the condition, as the system would not gain any insight through such sensing for this purpose. Besides, such informative channel can be sensed through the instalment of sensors, that can potentially result in encumbrance in the whole setup. Conversely, by sensing information about movements through time, the system is capable of inferring interaction forces with the user \unskip~\cite{ChenX2020Neural,Rozo2016a,Sasagawa2020}. This allows the robot to acquire a very crucial information to ensure safety in HRC, which is a key aspect for these applications.

On the same line, a minority of works makes use of human muscular activity as an input to the behavioural block of the system. However, it is primarily used jointly with vision \unskip~\cite{Zhou2019} or accelerometry \unskip~\cite{WangDiekel2018,Wang2018279}.

In some cases other sensing modalities were reported. Two cases of a sensor fusion system that uses human speech as a complementary modality to convey information in the robotic system are reported \unskip~\cite{WangDiekel2018,Wang2018279}. Moreover, Zhou et al. \unskip~\cite{Zhou2019} integrated user's brain activity in a sensor fusion composed by vision and muscular activity. Similarly to ML algorithms, only a few works make use of sensor fusion \unskip~\cite{Peternel2019,Roveda2019,WangDiekel2018,Wang2018279,Zhou2019}. Although their number is not that high to establish a pattern, it is interesting to point out that all the depicted modalities were combined in an almost equal number of cases, with no recurrent combination. 
    
\section{Discussions}
\label{discussions}
Every time dependent chart showed throughout this work (see Figures \ref{tasks_hist}, \ref{metrics_hist}, \ref{all_ml}, \ref{time_sensing} and Subfigure \ref{cognitive_time}) register a high increase of the number of papers regarding human-robot collaboration with the use of machine learning in 2018. This has to be attributed to the positioning paper of the International Federation of Robotics on the definition of cobot \unskip~\cite{IFR2019}. This is evidence that human-robot collaboration is taking progressively more importance in robotics research. Deployment of cobots in industrial settings is one of the main features in Industry 4.0 \unskip~\cite{Weiss2021}. So, research in this direction is encouraged. Besides, there are few works that make use of ML in the design of the architecture of the collaborative robot system compared to the amount of papers available regarding HRC. Therefore, works with such techniques surely are even more worth to be pursued. 

First, after having depicted features of the selected papers individually, more in-depth analysis of them is now carried out, to provide useful guidance for future works in this sector. A cross-analysis between the features themselves was performed. As comparing all possible pairings of them would result in an excessively lengthy dissertation, only the most relevant ones, in terms of results and meaningfulness, are reported. Finally, the last subsection deals with additional aspects related to HRC and gives suggestion for challenges to tackle in future research. 

In this section, other works not appeared as result of the selection process (see Section \ref{selection}) were used in this subsection, to enhance the discussion and give further contextualization to this review. 

\subsection{Machine learning for the design of collaborative tasks and cognitive variables}

In a general perspective, every work in the selection designs the experimental setup envisioning a human and a robot in it. None of them involves more than one subject in the setup at the same time (see Section \ref{experiments}). In human-robot interaction, over 150 works regarding non-dyadic interactions have been produced over the past 15 years \unskip~\cite{schneiders2022non}. So, considering them in HRC applications is highly encouraged.

Cross-analysing the use of ML techniques with the depicted categorization of collaborative tasks and cognitive variables allows to find interesting and useful correlations, that can be used by researchers as guidelines to design future works in HRC.

\begin{figure}[]
  \centering
  \includegraphics[width=\textwidth]{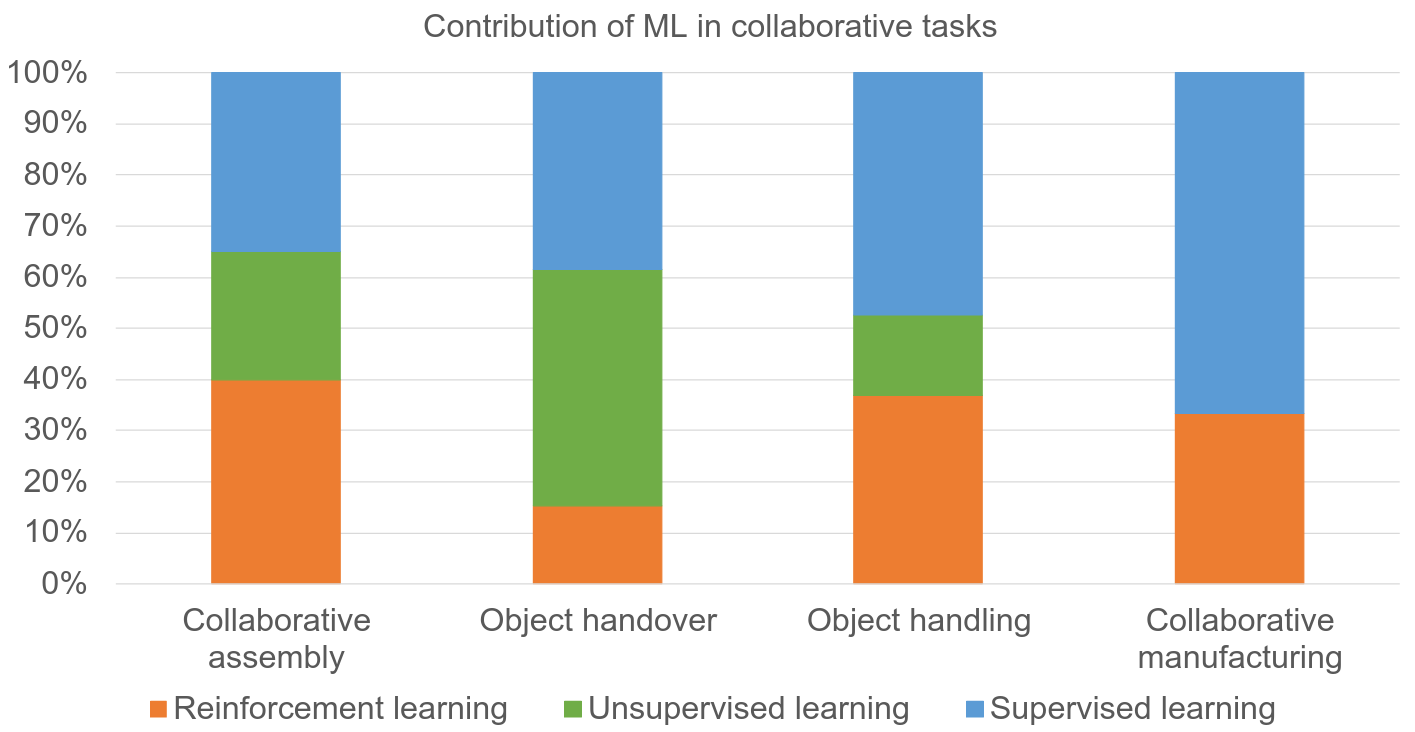}
  \hfill
  \caption{Comparison of the machine learning techniques exploited in the different categories of collaborative tasks. }
  \label{contribution_ml}
\end{figure}

Figure \ref{contribution_ml} provides insight about the use of ML techniques in the depicted categories of collaborative tasks (see Section \ref{collaborative}). 
Looking at collaborative assembly, all classes of ML algorithms are used, with a slight majority of RL algorithms. This result is to be expected: because of the design of the ML algorithm itself (see Section \ref{RL}), RL produces the policy itself for the robot to perform, that is convenient for a collaborative assembly task.

In object handling tasks, SL is the most used approach, followed by RL. This result carries an important message in terms of suggested research to pursue. Indeed, despite object handling being a task more related to robotic control, researchers are now employing ML techniques used to model more complex situations, like collaborative assembly, for this type of tasks, as well. This means that more complex and demanding versions of this task are being explored. Hence, using RL is advised for object handling tasks.

Regarding object handover, UL is predominant. This is because unsupervised learning is very used to model robot trajectories (see Subsection \ref{unsupervised}), which is very important in this type of task. So, these methodologies are suggested for this kind of works. Interestingly, there are very few cases of RL \unskip~\cite{Choi2018,Rozo2015}. This is because in object handover the interaction between user and robot is not continuous. Therefore, it is less relevant to employ a model more suited for scenarios where interactions are more time dependent (see Subsection \ref{RL}). 

Regarding collaborative manufacturing, there is a prevalence of supervised learning, followed by reinforcement learning, and no use of unsupervised learning. However, the low number of these tasks (see Figure \ref{tasks_hist}) does not allow to be certain of such a trend. However, for this fact itself, this type of application domains for HRC has high potential of discovery, dealing with still pretty untamed scenarios. 

\begin{figure}[H]
  \centering
  \includegraphics[width=\textwidth]{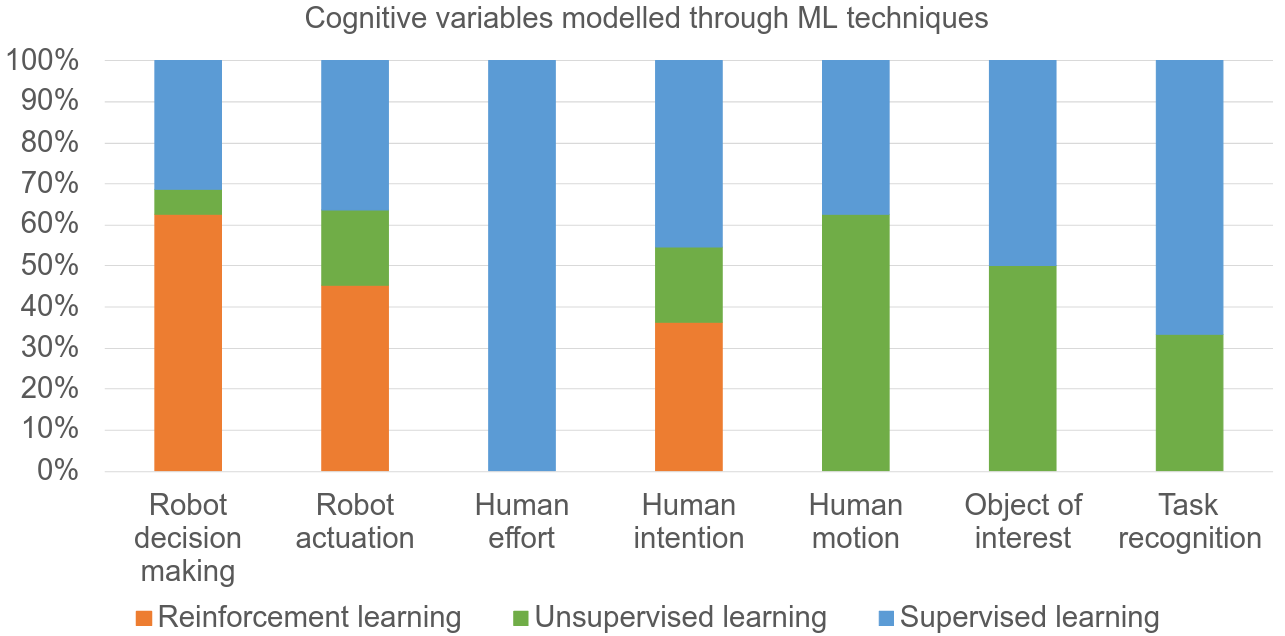}
  \hfill
  \caption{Frequency of occurrence of the ML techniques while modelling specific cognitive variables.}
  \label{ml_in_variables}
\end{figure}

To provide a further level of analysis for more detailed guidance on the use of ML in HRC, Figure \ref{ml_in_variables} shows what ML techniques are chosen when modelling cognitive system after certain variables. Regarding the cognitive variables related to the robot, it is possible to appreciate that RL is highly used to model them. This has again to be traced back to the implicit design of RL (see Subsection \ref{RL}). It produces the policy for an agent that collects observations from the environment, which is the ideal setting to model a robotic behaviour \unskip~\cite{Ghadirzadeh2020,Lu20201116,Roveda2020}.

About human-related cognitive variables, the situation varies, according to the specific variable. For human effort, there are too few cases to establish a pattern (see Subfigure \ref{pie_cognitive}). About human intention, there is no predominant methodology employed. Instead, for human motion, UL is the most used type of learning. This is again because UL models are proven to be good at modelling trajectories (see Subsection \ref{unsupervised}).

Regarding variables related to the scene, most of them regard the recognition of the task itself, with only two cases modelling the object of interest \unskip~\cite{Ahmad2020,Chi2018}. In these cases, SL is the most widely adopted approach, followed by UL. RL is not used, because it is more suitable to model output of the agent itself, rather than the environment (see Subsection \ref{RL}).

\subsection{Results in collaborative tasks}

\begin{figure}[H]
  \centering
  \includegraphics[width=\textwidth]{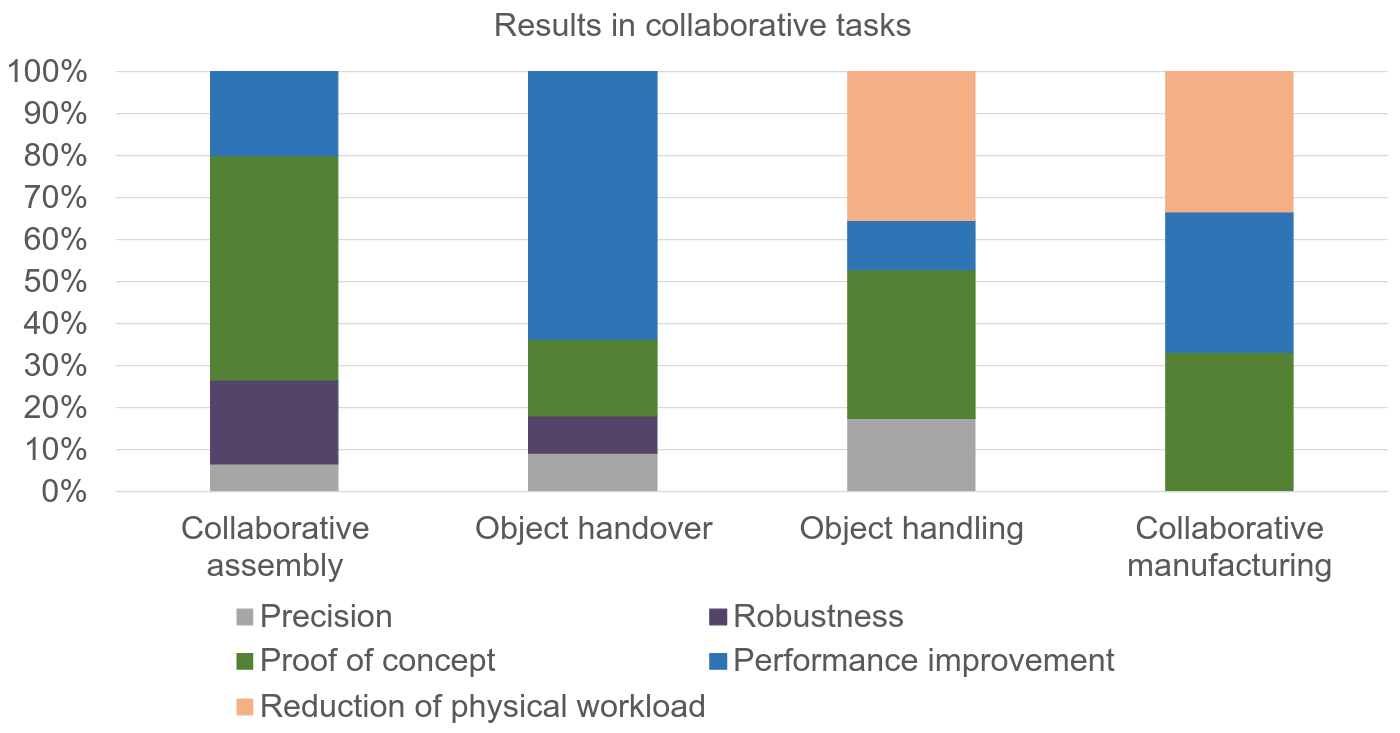}
  \hfill
  \caption{Frequency of occurrence of the categories of results achieved in the depicted collaborative tasks.}
  \label{results_collab}
\end{figure}

Exploring the classes of results for the selected categories allows to realise the depth of the research that has been reached and have a qualitative idea of the progress achieved so far in each of them (see Figure \ref{results_collab}). First, it is important to mention that every paper, when it applies, have measurements of success purposely defined to validate the efficiency of the robotic system itself. These variables are never  referred to more general standards. Therefore, basing HRC works results on derivations of standards, like NASA-TLX \unskip~\cite{story2022speed}, is suggested.

For collaborative assembly, some works already report results of a more specific kind, mainly about performance improvement and robustness. Anyway, the highest percentage of them still presents proof of concept as result. Research in collaborative assembly still has potential of discovery, although some standardised results are starting to be defined. Noticeably, there is a tendency to go more into robustness, as almost every work with this type of result belongs to collaborative assembly works. Indeed, in collaborative assembly, it is more crucial for cobots to be able to handle unexpected situations \unskip~\cite{Cunha2020}.

Figure \ref{results_collab} shows evidence of how much research in object handover tasks has reached a greater maturity. Indeed, almost every result related to this collaborative task deals with performance improvement with respect to a certain baseline condition. Indeed, this type of works, together with object handling tasks, have been going on longer than collaborative assembly \unskip~\cite{Matheson2019}. Similarly, in object handling the standard results relate to the reduction of the physical workload of the user during the task. Comparing this result with Figure \ref{ml_in_variables}, it is possible to make an interesting observation. The number of works that modelled human effort are much less than the works that have reduction of physical workload as goal. This implies that, interestingly, robotic researchers pursue such result without modelling the human effort directly, but by means of different cognitive variables.

About collaborative manufacturing, the number of works is not enough to work out a pattern (see Figure \ref{tasks_hist}), so more research in this direction should be pursued.

\subsection{Other perspectives}

This paper provides a systematic review of the use of machine learning techniques in human-robot collaboration. A combination of keywords was chosen to produce a set of unbiased results (see Section \ref{selection}). After this, the results were scanned and further selection criteria were designed to retain an adequate number of papers to perform a review about (see Section \ref{selection}). In doing so, other unfoldings of HRC and ML were excluded or did not appear. In this section, we briefly discuss about them and their potential for new research through the involvement of machine learning.

A subsector of ML that did not appear in the selection is transfer learning. Transfer learning makes use of ML algorithms of any kind to have a robotic system learn a certain skill under certain constraints of a ML problem, so that the same training model can be used for a different ML problem \unskip~\cite{LIU2020103515}. Ability to do so allows a robotic system to be versatile in HRC and be used under different settings than the ones in which it is validated \unskip~\cite{liu2019human}. Some works have recently addressed this problem by means of adversarial machine learning \unskip~\cite{LV2022108047}. Mainly related to DL models, core principle is having a ML system that learns the distribution of a dataset. Then, it generates a test set used by another ML system to understand whether a data point comes from the original distribution or the learnt one \unskip~\cite{yasar2021scalable}.  
Other ML techniques that can be used to perform transfer learning \unskip~\cite{LIU2020103515}, like inverse RL \unskip~\cite{WangDiekel2018} and deep RL \unskip~\cite{Ghadirzadeh2020}, appeared in the review, but were not used for such purpose. So, it is suggested to further explore this design paradigm in HRC. 

All the works reported in this review depict the use of a cobot in the experimental validation. Such architecture, in these works, is always represented by a low-payload robotic manipulator, from Sawyer \unskip~\cite{Cunha2020} and Baxter \unskip~\cite{ChenX2020Neural} to UR5 \unskip~\cite{Deng2017} and PR2 \unskip~\cite{Ghadirzadeh2016} and so on \unskip~\cite{Lorenzini2018,Nikolaidis2015,Wu2020Adaptive}. Instead, none of them envisions the use of a high-payload robot, like the COMAU NJ130 \unskip~\cite{Michalos2018}. However, these make up for a consistent amount of works in HRC \unskip~\cite{Michalos2022}. Indeed, robotics manufacturing was initially developed through the deployment of high-payload robots. They did not appear in the results of this review, because none of them uses ML techniques to develop the related case study \unskip~\cite{Dimitropoulos2021}. So, research with high-payload robots involving machine learning definitely is something to look into. Another robotic architecture falling in the categorization of HRC is robotic exoskeletons \unskip~\cite{Michalos2022}. There are plenty of works with such robotic architectures in HRC, but, similarly to high-payload robots, there are no works that merge advantages of robotic exoskeletons and ML together for HRC applications.

An approach very recently pursued in HRC is the use of digital twin for collaborative assembly, that is the realisation of a real-time digital counterpart that can re-enact the physical properties of the environment \unskip~\cite{Liu2019} and the manufacturing process. Here, collaborative workstations are simulated and the tasks are optimised, considering the presence of the worker in the scenario. A possible use of this approach is the optimization of trajectories of robotic manipulators in assembly lines \unskip~\cite{Bilberg2019}. Besides, two recent case studies have made use of machine learning techniques in a digital twin approach, namely neural networks \unskip~\cite{droder2018machine} and reinforcement learning \unskip~\cite{matulis2021robot}. This is evidence that researchers have only recently started to consider machine learning in this type of research, making this field very promising.

Kousi et al. \unskip~\cite{Kousi2021} also implemented a digital twin approach to optimise collaborative assembly, in which there were robots moving in the scene. Mobile robotics definitely brings much potential for applications in HRC \unskip~\cite{Michalos2022}. By being able to move in a factory and being endowed with manipulative capabilities \unskip~\cite{asfour2019armar}, robots could assist the worker in more dynamic tasks, that would not be constrained to only the space of the workstation. However, mobile manipulators seem not to be present jointly used with ML in HRC. If they do, the mobile ability of the robot is not used in the experimental validation (see Section \ref{experiments}). Hence, this other direction of research is suggested to be followed.

Finally, a great research thread in HRC is represented by safety \unskip~\cite{gualtieri2021emerging}. Many works are carried out evaluating feasibility of safe applications of collaborative robotics, but only few of them makes use of ML in it. For instance, Aivaliotis et al. \unskip~\cite{Aivaliotis2019} made use of neural networks to estimate the torque to increase the speed of the robot in reducing the amount of force exerted in simulated unexpected collisions between a human and a robot. Enabling safe collaborations by means of ML techniques is another research worth to pursue \unskip~\cite{Wang2019}.

\section{Conclusions}

\label{conclusions}
In this paper, a systematic review was performed, which led to the selection of 45 papers to find current trends in human-robot collaboration, with a specific focus on the use of machine learning techniques. 

A classification of different collaborative tasks was proposed, mainly composed by \textit{collaborative assembly}, \textit{object handover}, \textit{object handling} and \textit{collaborative manufacturing}. Alongside with it, metrics of these tasks were analysed. The cognitive variables modelled in the tasks were assessed and discussed, at different levels of details.

Machine learning techniques were subdivided in supervised, unsupervised and reinforcement learning, according to the learning algorithm used. The use of composite machine learning systems is not predominant in literature and it is encouraged, especially regarding deep reinforcement learning.

Regarding sensing, vision is the most used sensing modality in the selected set of papers, followed by accelerometry, with few works exploiting user's muscular activity. A very small portion of them uses other sensing modalities. Besides, similarly to the results concerning machine learning, sensor fusion is not much used in human-robot collaboration yet.

Further analysis was carried out, looking for correlations between the represented set of works' features. These were used to give guidelines to researchers for the design of future works that involve use of machine learning in human-robot collaboration applications. Among these, it was reported that reinforcement learning is predominantly used to model cognitive variables related to the robot itself, while unsupervised learning tends to be employed more in object handover and to model variables related to human motion. Looking at the types of results in the collaborative tasks, a good amount of them is still proof of concept, although some trends are starting to appear, like robustness for collaborative assembly and reduction of physical workload for object handling works.

Finally, aspects of human-robot collaboration research that did not fall under the scope of this review were mentioned, regarding their relationship with machine learning. The use of different robotic architectures than robotic manipulators, jointly with machine learning techniques, is suggested. Other domains and approaches that can benefit from machine learning are safety and use of digital twin for collaborative applications.

\section*{Declaration of competing interest}
The authors declare that they have no known competing financial interests or personal relationships that could have appeared to influence the work reported in this paper.

\section*{Acknowledgements}
Francesco Semeraro’s work was supported by an EPSRC CASE studentship funded by BAE Systems.
Angelo Cangelosi’s work was partially supported by the H2020 projects TRAINCREASE and eLADDA, the UKRI Trustworthy Autonomous Systems Node on Trust and the US Air Force project THRIVE++.

\bibliographystyle{elsarticle-num}

\bibliography{collection.bib}

\begin{thebibliography}{10}
\expandafter\ifx\csname url\endcsname\relax
  \def\url#1{\texttt{#1}}\fi
\expandafter\ifx\csname urlprefix\endcsname\relax\def\urlprefix{URL }\fi
\expandafter\ifx\csname href\endcsname\relax
  \def\href#1#2{#2} \def\path#1{#1}\fi

\bibitem{IFR2019}
{A positioning paper by the International Federation of Robotics WHAT IS A
  COLLABORATIVE INDUSTRIAL ROBOT?} (2018).

\bibitem{bauer2008human}
A.~Bauer, D.~Wollherr, M.~Buss, Human--robot collaboration: a survey,
  International Journal of Humanoid Robotics 5 (2008) 47--66.

\bibitem{Matheson2019}
E.~Matheson, R.~Minto, E.~G.~G. Zampieri, M.~Faccio, G.~Rosati, {Human–Robot
  Collaboration in Manufacturing Applications: A Review}, MDPI Robotics 8
  (2019) 100.
\newblock \href {https://doi.org/10.3390/robotics8040100}
  {\path{doi:10.3390/robotics8040100}}.

\bibitem{ISI:000389809100162}
K.~E. Kaplan, {Improving Inclusion Segmentation Task Performance Through
  Human-Intent Based Human-Robot Collaboration}, 11th ACM IEEE International
  Conference on Human-Robot Interaction, 2016, pp. 623--624.

\bibitem{jacob2012gestonurse}
M.~Jacob, Y.-T. Li, G.~Akingba, J.~P. Wachs, Gestonurse: a robotic surgical
  nurse for handling surgical instruments in the operating room, Journal of
  Robotic Surgery 6 (2012) 53--63.

\bibitem{amarillo2021collaborative}
A.~Amarillo, E.~Sanchez, J.~Caceres, J.~O{\~n}ativia, Collaborative
  human--robot interaction interface: Development for a spinal surgery robotic
  assistant, International Journal of Social Robotics (2021) 1--12.

\bibitem{Ogata2022}
T.~Ogata, K.~Takahashi, T.~Yamada, S.~Murata, K.~Sasaki, Machine learning for
  cognitive robotics, Cognitive Robotics (2022).

\bibitem{EmeohaOgenyi}
U.~{Emeoha Ogenyi}, S.~Member, J.~Liu, S.~Member, C.~Yang, Z.~Ju, H.~Liu,
  {Physical Human-Robot Collaboration: Robotic Systems, Learning Methods,
  Collaborative Strategies, Sensors and Actuators}, Tech. rep.

\bibitem{Hiatt2017}
L.~M. Hiatt, C.~Narber, E.~Bekele, S.~S. Khemlani, J.~G. Trafton, {Human
  modeling for human–robot collaboration}, The International Journal of
  Robotics Research 36 (2017) 580--596.
\newblock \href {https://doi.org/10.1177/0278364917690592}
  {\path{doi:10.1177/0278364917690592}}.

\bibitem{Chandrasekaran2015}
B.~Chandrasekaran, J.~M. Conrad, {Human-robot collaboration: A survey}, in:
  Conference Proceedings - IEEE SOUTHEASTCON, Vol. 2015-June, 2015.
\newblock \href {https://doi.org/10.1109/SECON.2015.7132964}
  {\path{doi:10.1109/SECON.2015.7132964}}.

\bibitem{mukherjee2022survey}
D.~Mukherjee, K.~Gupta, L.~H. Chang, H.~Najjaran, A survey of robot learning
  strategies for human-robot collaboration in industrial settings, Robotics and
  Computer-Integrated Manufacturing 73 (2022) 102231.

\bibitem{Ahmad2020}
M.~A. Ahmad, M.~Ourak, C.~Gruijthuijsen, J.~Deprest, T.~Vercauteren, E.~{Vander
  Poorten}, {Deep learning-based monocular placental pose estimation: towards
  collaborative robotics in fetoscopy}, International Journal of Computer
  Assisted Radiology and Surgery 15 (2020) 1561--1571.
\newblock \href {https://doi.org/10.1007/s11548-020-02166-3}
  {\path{doi:10.1007/s11548-020-02166-3}}.

\bibitem{Akkaladevi20191491}
S.~C. Akkaladevi, M.~Plasch, A.~Pichler, M.~Ikeda, Towards reinforcement based
  learning of an assembly process for human robot collaboration, in: Procedia
  Manufacturing, Vol.~38, 2019, pp. 1491--1498.
\newblock \href {https://doi.org/10.1016/j.promfg.2020.01.138}
  {\path{doi:10.1016/j.promfg.2020.01.138}}.

\bibitem{ChenM2020}
M.~Chen, H.~Soh, D.~Hsu, S.~Nikolaidis, S.~Srinivasa, {Trust-aware decision
  making for human-robot collaboration: Model learning and planning}, ACM
  Transactions on Human-Robot Interaction 9 (jan 2020).
\newblock \href {http://arxiv.org/abs/1801.04099} {\path{arXiv:1801.04099}},
  \href {https://doi.org/10.1145/3359616} {\path{doi:10.1145/3359616}}.

\bibitem{ChenX2020stiffness}
X.~Chen, Y.~Jiang, C.~Yang, {Stiffness Estimation and Intention Detection for
  Human-Robot Collaboration}, in: Proceedings of the 15th IEEE Conference on
  Industrial Electronics and Applications, ICIEA 2020, 2020, pp. 1802--1807.
\newblock \href {https://doi.org/10.1109/ICIEA48937.2020.9248186}
  {\path{doi:10.1109/ICIEA48937.2020.9248186}}.

\bibitem{ChenX2020Neural}
X.~Chen, N.~Wang, H.~Cheng, C.~Yang, {Neural Learning Enhanced Variable
  Admittance Control for Human-Robot Collaboration}, IEEE Access 8 (2020)
  25727--25737.
\newblock \href {https://doi.org/10.1109/ACCESS.2020.2969085}
  {\path{doi:10.1109/ACCESS.2020.2969085}}.

\bibitem{Chi2018}
W.~Chi, J.~Liu, H.~Rafii-Tari, C.~Riga, C.~Bicknell, G.~Z. Yang,
  {Learning-based endovascular navigation through the use of non-rigid
  registration for collaborative robotic catheterization}, International
  Journal of Computer Assisted Radiology and Surgery 13 (2018) 855--864.
\newblock \href {https://doi.org/10.1007/s11548-018-1743-5}
  {\path{doi:10.1007/s11548-018-1743-5}}.

\bibitem{Choi2018}
S.~Choi, K.~Lee, H.~A. Park, S.~Oh, {A Nonparametric Motion Flow Model for
  Human Robot Cooperation}, IEEE International Conference on Robotics and
  Automation ICRA, 2018, pp. 7211--7218.

\bibitem{Chung2015}
M.~J.~Y. Chung, A.~L. Friesen, D.~Fox, A.~N. Meltzoff, R.~P. Rao, {A Bayesian
  developmental approach to robotic goal-based imitation learning}, PLoS ONE 10
  (nov 2015).
\newblock \href {https://doi.org/10.1371/journal.pone.0141965}
  {\path{doi:10.1371/journal.pone.0141965}}.

\bibitem{Cunha2020}
A.~Cunha, F.~Ferreira, E.~Sousa, L.~Louro, P.~Vicente, S.~Monteiro,
  W.~Erlhagen, E.~Bicho, {Towards Collaborative Robots as Intelligent
  Co-workers in Human-Robot Joint Tasks: what to do and who does it?}, in: ISR
  2020; 52th International Symposium on Robotics, 2020, pp. 1--8.

\bibitem{Deng2017}
Z.~Deng, J.~Mi, D.~Han, R.~Huang, X.~Xiong, J.~Zhang, {Hierarchical Robot
  Learning for Physical Collaboration between Humans and Robots}, in: 2017 IEEE
  International Conference on Robotics and Biomimetics (IEEE ROBIO 2017), 2017,
  pp. 750--755.

\bibitem{Ghadirzadeh2016}
A.~Ghadirzadeh, J.~B{\"{u}}tepage, A.~Maki, D.~Kragic, M.~Bj{\"{o}}rkman, {A
  sensorimotor reinforcement learning framework for physical human-robot
  interaction}, in: IEEE International Conference on Intelligent Robots and
  Systems, 2016, pp. 2682--2688.
\newblock \href {http://arxiv.org/abs/1607.07939} {\path{arXiv:1607.07939}},
  \href {https://doi.org/10.1109/IROS.2016.7759417}
  {\path{doi:10.1109/IROS.2016.7759417}}.

\bibitem{Ghadirzadeh2020}
A.~Ghadirzadeh, X.~Chen, W.~Yin, Z.~Yi, M.~Bj{\"{o}}rkman, D.~Kragic,
  M.~Bjorkman, D.~Kragic, {Human-centered collaborative robots with deep
  reinforcement learning}, IEEE Robotics and Automation Letters 6 (2020)
  566--571.
\newblock \href {https://doi.org/10.1109/LRA.2020.3047730}
  {\path{doi:10.1109/LRA.2020.3047730}}.

\bibitem{Grigore2018}
E.~C. Grigore, A.~Roncone, O.~Mangin, B.~Scassellati, {Preference-Based
  Assistance Prediction for Human-Robot Collaboration Tasks}, in: IEEE
  International Conference on Intelligent Robots and Systems, 2018, pp.
  4441--4448.
\newblock \href {https://doi.org/10.1109/IROS.2018.8593716}
  {\path{doi:10.1109/IROS.2018.8593716}}.

\bibitem{Huang2018}
Y.~Huang, J.~Silverio, L.~Rozo, D.~G. Caldwell, {Generalized Task-Parameterized
  Skill Learning}, IEEE International Conference on Robotics and Automation
  ICRA, 2018, pp. 5667--5674.

\bibitem{Kuang2018}
S.~Kuang, Y.~Tang, A.~Lin, S.~Yu, L.~Sun, {Intelligent Control for Human-Robot
  Cooperation in Orthopedics Surgery}, Vol. 1093 of Advances in Experimental
  Medicine and Biology, 2018, pp. 245--262.
\newblock \href {https://doi.org/10.1007/978-981-13-1396-7_19}
  {\path{doi:10.1007/978-981-13-1396-7_19}}.

\bibitem{Lorenzini2018}
M.~Lorenzini, W.~Kim, E.~{De Momi}, A.~Ajoudani, {A synergistic approach to the
  Real-Time estimation of the feet ground reaction forces and centers of
  pressure in humans with application to Human-Robot collaboration}, IEEE
  Robotics and Automation Letters 3 (2018) 3654--3661.
\newblock \href {https://doi.org/10.1109/LRA.2018.2855802}
  {\path{doi:10.1109/LRA.2018.2855802}}.

\bibitem{Lu20201116}
W.~Lu, Z.~Hu, J.~Pan, {Human-Robot Collaboration using Variable Admittance
  Control and Human Intention Prediction}, in: IEEE International Conference on
  Automation Science and Engineering, Vol. 2020-Augus, 2020, pp. 1116--1121.
\newblock \href {https://doi.org/10.1109/CASE48305.2020.9217040}
  {\path{doi:10.1109/CASE48305.2020.9217040}}.

\bibitem{Luo2018}
R.~Luo, R.~Hayne, D.~Berenson, {Unsupervised early prediction of human reaching
  for human–robot collaboration in shared workspaces}, Autonomous Robots 42
  (2018) 631--648.
\newblock \href {https://doi.org/10.1007/s10514-017-9655-8}
  {\path{doi:10.1007/s10514-017-9655-8}}.

\bibitem{MaedaNeumann2017}
G.~J. Maeda, G.~Neumann, M.~Ewerton, R.~Lioutikov, O.~Kroemer, J.~Peters,
  {Probabilistic movement primitives for coordination of multiple human–robot
  collaborative tasks}, Autonomous Robots 41 (2017) 593--612.
\newblock \href {https://doi.org/10.1007/s10514-016-9556-2}
  {\path{doi:10.1007/s10514-016-9556-2}}.

\bibitem{MaedaEwerton2017}
G.~Maeda, M.~Ewerton, G.~Neumann, R.~Lioutikov, J.~Peters, {Phase estimation
  for fast action recognition and trajectory generation in human–robot
  collaboration}, The International Journal of Robotics Research 36 (2017)
  1579--1594.
\newblock \href {https://doi.org/10.1177/0278364917693927}
  {\path{doi:10.1177/0278364917693927}}.

\bibitem{Munzer2018}
T.~Munzer, M.~Toussaint, M.~Lopes, {Efficient behavior learning in
  human–robot collaboration}, Autonomous Robots 42 (2018) 1103--1115.
\newblock \href {https://doi.org/10.1007/s10514-017-9674-5}
  {\path{doi:10.1007/s10514-017-9674-5}}.

\bibitem{MurataLi2018}
S.~Murata, Y.~Li, H.~Arie, T.~Ogata, S.~Sugano, {Learning to Achieve Different
  Levels of Adaptability for Human-Robot Collaboration Utilizing a
  Neuro-Dynamical System}, IEEE Transactions on Cognitive and Developmental
  Systems 10 (2018) 712--725.
\newblock \href {https://doi.org/10.1109/TCDS.2018.2797260}
  {\path{doi:10.1109/TCDS.2018.2797260}}.

\bibitem{MurataMasuda2019}
S.~Murata, W.~Masuda, J.~Chen, H.~Arie, T.~Ogata, S.~Sugano, {Achieving
  Human–Robot Collaboration with Dynamic Goal Inference by Gradient Descent},
  Lecture Notes in Computer Science (including subseries Lecture Notes in
  Artificial Intelligence and Lecture Notes in Bioinformatics) 11954 LNCS
  (2019) 579--590.
\newblock \href {https://doi.org/10.1007/978-3-030-36711-4_49}
  {\path{doi:10.1007/978-3-030-36711-4_49}}.

\bibitem{Nikolaidis2015}
S.~Nikolaidis, R.~Ramakrishnan, K.~Gu, J.~Shah, {Efficient Model Learning from
  Joint-Action Demonstrations for Human-Robot Collaborative Tasks}, ACM/IEEE
  International Conference on Human-Robot Interaction 2015-March (2015)
  189--196.
\newblock \href {https://doi.org/10.1145/2696454.2696455}
  {\path{doi:10.1145/2696454.2696455}}.

\bibitem{Peternel2019}
L.~Peternel, C.~Fang, N.~Tsagarakis, A.~Ajoudani, {A selective muscle fatigue
  management approach to ergonomic human-robot co-manipulation}, Robotics and
  Computer-Integrated Manufacturing 58 (2019) 69--79.
\newblock \href {https://doi.org/10.1016/j.rcim.2019.01.013}
  {\path{doi:10.1016/j.rcim.2019.01.013}}.

\bibitem{Roveda2019}
L.~Roveda, S.~Haghshenas, M.~Caimmi, N.~Pedrocchi, L.~M. Tosatti, {Assisting
  operators in heavy industrial tasks: On the design of an optimized
  cooperative impedance fuzzy-controller with embedded safety rules}, Frontiers
  in Robotics and AI 6 (aug 2019).
\newblock \href {https://doi.org/10.3389/frobt.2019.00075}
  {\path{doi:10.3389/frobt.2019.00075}}.

\bibitem{Roveda2020}
L.~Roveda, J.~Maskani, P.~Franceschi, A.~Abdi, F.~Braghin, L.~{Molinari
  Tosatti}, N.~Pedrocchi, {Model-Based Reinforcement Learning Variable
  Impedance Control for Human-Robot Collaboration}, Journal of Intelligent and
  Robotic Systems: Theory and Applications 100 (2020) 417--433.
\newblock \href {https://doi.org/10.1007/s10846-020-01183-3}
  {\path{doi:10.1007/s10846-020-01183-3}}.

\bibitem{Rozo2015}
L.~Rozo, D.~Bruno, S.~Calinon, D.~G. Caldwell, {Learning Optimal Controllers in
  Human-robot Cooperative Transportation Tasks with Position and Force
  Constraints}, in: 2015 IEEE/RSJ International Conference on Intelligent
  Robots and Systems (IROS), 2015, pp. 1024--1030.

\bibitem{Rozo2016a}
L.~Rozo, J.~Silv{\'{e}}rio, S.~Calinon, D.~G. Caldwell, {Learning controllers
  for reactive and proactive behaviors in human-robot collaboration}, Frontiers
  Robotics AI 3 (jun 2016).
\newblock \href {https://doi.org/10.3389/frobt.2016.00030}
  {\path{doi:10.3389/frobt.2016.00030}}.

\bibitem{Sasagawa2020}
A.~Sasagawa, K.~Fujimoto, S.~Sakaino, T.~Tsuji, {Imitation Learning Based on
  Bilateral Control for Human-Robot Cooperation}, IEEE Robotics and Automation
  Letters 5 (2020) 6169--6176.
\newblock \href {https://doi.org/10.1109/LRA.2020.3011353}
  {\path{doi:10.1109/LRA.2020.3011353}}.

\bibitem{Shukla2018}
D.~Shukla, O.~Erkent, J.~Piater, {Learning semantics of gestural instructions
  for human-robot collaboration}, Frontiers in Neurorobotics 12 (mar 2018).
\newblock \href {https://doi.org/10.3389/fnbot.2018.00007}
  {\path{doi:10.3389/fnbot.2018.00007}}.

\bibitem{Tabrez2019}
A.~Tabrez, S.~Agrawal, B.~Hayes, {Explanation-based Reward Coaching to Improve
  Human Performance via Reinforcement Learning}, ACM IEEE International
  Conference on Human-Robot Interaction, 2019, pp. 249--257.

\bibitem{Tsiakas2017423}
K.~Tsiakas, M.~Papakostas, M.~Papakostas, M.~Bell, R.~Mihalcea, S.~Wang,
  M.~Burzo, F.~Makedon, {An interactive multisensing framework for personalized
  human robot collaboration and assistive training using reinforcement
  learning}, in: ACM International Conference Proceeding Series, Vol. Part
  F1285, 2017, pp. 423--427.
\newblock \href {https://doi.org/10.1145/3056540.3076191}
  {\path{doi:10.1145/3056540.3076191}}.

\bibitem{Unhelkar2020}
V.~V. Unhelkar, S.~Li, J.~A. Shah, {Decision-making for bidirectional
  communication in sequential human-robot collaborative tasks}, in: ACM/IEEE
  International Conference on Human-Robot Interaction, New York, NY, USA, 2020,
  pp. 329--341.
\newblock \href {https://doi.org/10.1145/3319502.3374779}
  {\path{doi:10.1145/3319502.3374779}}.

\bibitem{Spaa9197296}
L.~v.~der Spaa, M.~Gienger, T.~Bates, J.~Kober, {Predicting and Optimizing
  Ergonomics in Physical Human-Robot Cooperation Tasks}, in: 2020 IEEE
  International Conference on Robotics and Automation (ICRA), 2020, pp.
  1799--1805.
\newblock \href {https://doi.org/10.1109/ICRA40945.2020.9197296}
  {\path{doi:10.1109/ICRA40945.2020.9197296}}.

\bibitem{Vinanzi2020}
S.~Vinanzi, A.~Cangelosi, C.~Goerick, {The Role of Social Cues for Goal
  Disambiguation in Human-Robot Cooperation}, in: 2020 29th IEEE International
  Conference on Robot and Human Interactive Communication (RO-MAN), 2020, pp.
  971--977.
\newblock \href {https://doi.org/10.1109/RO-MAN47096.2020.9223546}
  {\path{doi:10.1109/RO-MAN47096.2020.9223546}}.

\bibitem{Vogt2016}
D.~Vogt, S.~Stepputtis, R.~Weinhold, B.~Jung, H.~{Ben Amor}, {Learning
  Human-Robot Interactions from Human-Human Demonstrations (with Applications
  in Lego Rocket Assembly)}, IEEE-RAS International Conference on Humanoid
  Robots, 2016, pp. 142--143.

\bibitem{WangDiekel2018}
W.~Wang, R.~Li, Y.~Chen, Z.~M. Diekel, Y.~Jia, Facilitating human--robot
  collaborative tasks by teaching-learning-collaboration from human
  demonstrations, IEEE Transactions on Automation Science and Engineering 16
  (2018) 640--653.

\bibitem{Wang2018279}
W.~Wang, R.~Li, Y.~Chen, Y.~Jia, {Human Intention Prediction in Human-Robot
  Collaborative Tasks}, in: ACM/IEEE International Conference on Human-Robot
  Interaction, 2018, pp. 279--280.
\newblock \href {https://doi.org/10.1145/3173386.3177025}
  {\path{doi:10.1145/3173386.3177025}}.

\bibitem{Wojtak}
W.~Wojtak, F.~Ferreira, P.~Vicente, L.~Louro, E.~Bicho, W.~Erlhagen, {A neural
  integrator model for planning and value-based decision making of a robotics
  assistant}, Neural Computing \& Applications (2020).
\newblock \href {https://doi.org/10.1007/s00521-020-05224-8}
  {\path{doi:10.1007/s00521-020-05224-8}}.

\bibitem{Wu2020Adaptive}
M.~Wu, Y.~He, S.~Liu, {Adaptive impedance control based on reinforcement
  learning in a human-robot collaboration task with human reference
  estimation}, International Journal of Mechanics and Control 21 (2020) 21--31.

\bibitem{Wu2020Shared}
M.~Wu, Y.~He, S.~Liu, {Shared Impedance Control Based on Reinforcement Learning
  in a Human-Robot Collaboration Task}, Vol. 980 of Advances in Intelligent
  Systems and Computing, 2020, pp. 95--103.
\newblock \href {https://doi.org/10.1007/978-3-030-19648-6_12}
  {\path{doi:10.1007/978-3-030-19648-6_12}}.

\bibitem{Yan20191390}
L.~Yan, X.~Gao, X.~Zhang, S.~Chang, {Human-robot collaboration by intention
  recognition using deep LSTM neural network}, in: Proceedings of the 8th
  International Conference on Fluid Power and Mechatronics, FPM 2019, 2019, pp.
  1390--1396.
\newblock \href {https://doi.org/10.1109/FPM45753.2019.9035907}
  {\path{doi:10.1109/FPM45753.2019.9035907}}.

\bibitem{Zhang2020}
J.~Zhang, H.~Liu, Q.~Chang, L.~Wang, R.~X. Gao, {Recurrent neural network for
  motion trajectory prediction in human-robot collaborative assembly}, CIRP
  Annals 69 (2020) 9--12.
\newblock \href {https://doi.org/10.1016/j.cirp.2020.04.077}
  {\path{doi:10.1016/j.cirp.2020.04.077}}.

\bibitem{Zhou2019}
T.~Zhou, J.~P. Wachs, {Spiking Neural Networks for early prediction in
  human–robot collaboration}, International Journal of Robotics Research 38
  (2019) 1619--1643.
\newblock \href {http://arxiv.org/abs/1807.11096} {\path{arXiv:1807.11096}},
  \href {https://doi.org/10.1177/0278364919872252}
  {\path{doi:10.1177/0278364919872252}}.

\bibitem{jo2021machine}
T.~Jo, Machine Learning Foundations: Supervised, Unsupervised, and Advanced
  Learning, Springer Nature, 2021.

\bibitem{Weiss2021}
A.~Weiss, A.~K. Wortmeier, B.~Kubicek, {Cobots in Industry 4.0: A Roadmap for
  Future Practice Studies on Human-Robot Collaboration}, IEEE Transactions on
  Human-Machine Systems 51~(4) (2021) 335--345.
\newblock \href {https://doi.org/10.1109/THMS.2021.3092684}
  {\path{doi:10.1109/THMS.2021.3092684}}.

\bibitem{schneiders2022non}
E.~Schneiders, E.~Cheon, J.~Kjeldskov, M.~Rehm, M.~B. Skov, Non-dyadic
  interaction: A literature review of 15 years of human-robot interaction
  conference publications, ACM Transactions on Human-Robot Interaction (THRI)
  11~(2) (2022) 1--32.

\bibitem{story2022speed}
M.~Story, P.~Webb, S.~R. Fletcher, G.~Tang, C.~Jaksic, J.~Carberry, Do speed
  and proximity affect human-robot collaboration with an industrial robot arm?,
  International Journal of Social Robotics (2022) 1--16.

\bibitem{LIU2020103515}
Y.~Liu, Z.~Li, H.~Liu, Z.~Kan,
  \href{https://www.sciencedirect.com/science/article/pii/S0921889019309972}{Skill
  transfer learning for autonomous robots and human–robot cooperation: A
  survey}, Robotics and Autonomous Systems 128 (2020) 103515.
\newblock \href {https://doi.org/https://doi.org/10.1016/j.robot.2020.103515}
  {\path{doi:https://doi.org/10.1016/j.robot.2020.103515}}.
\newline\urlprefix\url{https://www.sciencedirect.com/science/article/pii/S0921889019309972}

\bibitem{liu2019human}
Q.~Liu, Z.~Liu, W.~Xu, Q.~Tang, Z.~Zhou, D.~T. Pham, Human-robot collaboration
  in disassembly for sustainable manufacturing, International Journal of
  Production Research 57~(12) (2019) 4027--4044.

\bibitem{LV2022108047}
Q.~Lv, R.~Zhang, T.~Liu, P.~Zheng, Y.~Jiang, J.~Li, J.~Bao, L.~Xiao,
  \href{https://www.sciencedirect.com/science/article/pii/S0360835222001176}{A
  strategy transfer approach for intelligent human-robot collaborative
  assembly}, Computers \& Industrial Engineering 168 (2022) 108047.
\newblock \href {https://doi.org/https://doi.org/10.1016/j.cie.2022.108047}
  {\path{doi:https://doi.org/10.1016/j.cie.2022.108047}}.
\newline\urlprefix\url{https://www.sciencedirect.com/science/article/pii/S0360835222001176}

\bibitem{yasar2021scalable}
M.~S. Yasar, T.~Iqbal, A scalable approach to predict multi-agent motion for
  human-robot collaboration, IEEE Robotics and Automation Letters 6~(2) (2021)
  1686--1693.

\bibitem{Michalos2018}
G.~Michalos, N.~Kousi, P.~Karagiannis, C.~Gkournelos, K.~Dimoulas, S.~Koukas,
  K.~Mparis, A.~Papavasileiou, S.~Makris, {Seamless human robot collaborative
  assembly – An automotive case study}, Mechatronics 55 (2018) 194--211.
\newblock \href {https://doi.org/10.1016/J.MECHATRONICS.2018.08.006}
  {\path{doi:10.1016/J.MECHATRONICS.2018.08.006}}.

\bibitem{Michalos2022}
G.~Michalos, P.~Karagiannis, N.~Dimitropoulos, D.~Andronas, S.~Makris, {Human
  Robot Collaboration in Industrial Environments}, Intelligent Systems, Control
  and Automation: Science and Engineering 81 (2022) 17--39.
\newblock \href {https://doi.org/10.1007/978-3-030-78513-0_2}
  {\path{doi:10.1007/978-3-030-78513-0_2}}.

\bibitem{Dimitropoulos2021}
N.~Dimitropoulos, T.~Togias, N.~Zacharaki, G.~Michalos, S.~Makris, {Seamless
  Human–Robot Collaborative Assembly Using Artificial Intelligence and
  Wearable Devices}, Applied Sciences 2021, Vol. 11, Page 5699 11~(12) (2021)
  5699.
\newblock \href {https://doi.org/10.3390/APP11125699}
  {\path{doi:10.3390/APP11125699}}.

\bibitem{Liu2019}
Q.~Liu, Z.~Liu, W.~Xu, Q.~Tang, Z.~Zhou, D.~T. Pham, {Human-robot collaboration
  in disassembly for sustainable manufacturing}, International Journal of
  Production Research 57 (2019) 4027--4044.
\newblock \href {https://doi.org/10.1080/00207543.2019.1578906}
  {\path{doi:10.1080/00207543.2019.1578906}}.

\bibitem{Bilberg2019}
A.~Bilberg, A.~A. Malik, {Digital twin driven human–robot collaborative
  assembly}, CIRP Annals 68~(1) (2019) 499--502.
\newblock \href {https://doi.org/10.1016/J.CIRP.2019.04.011}
  {\path{doi:10.1016/J.CIRP.2019.04.011}}.

\bibitem{droder2018machine}
K.~Dr{\"o}der, P.~Bobka, T.~Germann, F.~Gabriel, F.~Dietrich, A machine
  learning-enhanced digital twin approach for human-robot-collaboration,
  Procedia Cirp 76 (2018) 187--192.

\bibitem{matulis2021robot}
M.~Matulis, C.~Harvey, A robot arm digital twin utilising reinforcement
  learning, Computers \& Graphics 95 (2021) 106--114.

\bibitem{Kousi2021}
N.~Kousi, C.~Gkournelos, S.~Aivaliotis, K.~Lotsaris, A.~C. Bavelos, P.~Baris,
  G.~Michalos, S.~Makris, {Digital Twin for Designing and Reconfiguring
  Human–Robot Collaborative Assembly Lines}, Applied Sciences 2021, Vol. 11,
  Page 4620 11~(10) (2021) 4620.
\newblock \href {https://doi.org/10.3390/APP11104620}
  {\path{doi:10.3390/APP11104620}}.

\bibitem{asfour2019armar}
T.~Asfour, M.~Waechter, L.~Kaul, S.~Rader, P.~Weiner, S.~Ottenhaus, R.~Grimm,
  Y.~Zhou, M.~Grotz, F.~Paus, Armar-6: A high-performance humanoid for
  human-robot collaboration in real-world scenarios, IEEE Robotics \&
  Automation Magazine 26~(4) (2019) 108--121.

\bibitem{gualtieri2021emerging}
L.~Gualtieri, E.~Rauch, R.~Vidoni, Emerging research fields in safety and
  ergonomics in industrial collaborative robotics: A systematic literature
  review, Robotics and Computer-Integrated Manufacturing 67 (2021) 101998.

\bibitem{Aivaliotis2019}
P.~Aivaliotis, S.~Aivaliotis, C.~Gkournelos, K.~Kokkalis, G.~Michalos,
  S.~Makris, {Power and force limiting on industrial robots for human-robot
  collaboration}, Robotics and Computer-Integrated Manufacturing 59 (2019)
  346--360.
\newblock \href {https://doi.org/10.1016/J.RCIM.2019.05.001}
  {\path{doi:10.1016/J.RCIM.2019.05.001}}.

\bibitem{Wang2019}
L.~Wang, R.~Gao, J.~V{\'{a}}ncza, J.~Kr{\"{u}}ger, X.~V. Wang, S.~Makris,
  G.~Chryssolouris, {Symbiotic human-robot collaborative assembly}, CIRP Annals
  68~(2) (2019) 701--726.
\newblock \href {https://doi.org/10.1016/J.CIRP.2019.05.002}
  {\path{doi:10.1016/J.CIRP.2019.05.002}}.

\end{thebibliography}


\end{document}